\documentclass[10pt,twocolumn,letterpaper]{article}

\usepackage[pagenumbers]{cvpr}

\def\confName{CVPR}
\def\confYear{2027}
\usepackage{microtype}
\usepackage{amsfonts}
\usepackage{multirow}
\usepackage{array}
\usepackage{tabularx}
\usepackage{colortbl}
\usepackage{algorithm}
\usepackage{algpseudocode}
\usepackage{tikz}
\usetikzlibrary{arrows.meta,positioning,calc,fit,backgrounds}
\definecolor{cvprblue}{rgb}{0.21,0.49,0.74}
\usepackage[pagebackref,breaklinks,colorlinks,allcolors=cvprblue]{hyperref}
\usepackage[capitalize,noabbrev]{cleveref}
\newcommand{\method}{TEC\xspace}
\newcommand{\pruner}{CoverFold\xspace}

\newcommand{\ci}[3]{#1\,{\scriptsize[#2,#3]}}

\newcommand{\Rset}{\mathcal{R}}
\newcommand{\Dset}{\mathcal{D}}
\newcommand{\Sset}{\mathcal{S}}
\newcommand{\Pset}{\mathcal{P}}
\definecolor{sourceorange}{HTML}{CF6F24}
\definecolor{memoryblue}{HTML}{25658E}
\definecolor{writegreen}{HTML}{2C7C60}
\definecolor{mutedgray}{HTML}{5E6870}
\newcolumntype{Y}{>{\raggedright\arraybackslash}X}

\title{From Overhead to Memory: Consolidating Visual Evidence into Timestamps\\for Temporal-Aware Token Compression in Streaming Video LLMs}
\author{%
  Jiaqi Li\textsuperscript{1},
  Guangming Wang\textsuperscript{2},
  Shuntian Zheng\textsuperscript{1},
  Xiaoman Lu\textsuperscript{1},
  Hoang Hai Pham\textsuperscript{1},
  Shuyan Zhang\textsuperscript{1},\\
  Yue Wen\textsuperscript{3},
  Yixian Shen\textsuperscript{4},
  Yongbo Chen\textsuperscript{3},
  Sai Gu\textsuperscript{1},
  Hesheng Wang\textsuperscript{3},
  Yu Guan\textsuperscript{1}\\[3pt]
  {\normalsize \textsuperscript{1}The University of Warwick\quad
  \textsuperscript{2}University of Cambridge}\\
  {\normalsize \textsuperscript{3}Shanghai Jiao Tong University\quad
  \textsuperscript{4}University of Amsterdam}%
}
\hypersetup{pdfauthor={Jiaqi Li, Guangming Wang, Shuntian Zheng, Xiaoman Lu,
  Hoang Hai Pham, Shuyan Zhang, Yue Wen, Yixian Shen, Yongbo Chen, Sai Gu,
  Hesheng Wang, Yu Guan}}

\begin{document}
\maketitle
\begin{abstract}
Streaming video Large Language Models (LLMs) attend to growing visual histories, increasing Key--Value (KV) cache  and per-frame attention cost with video length and threatening real-time processing.
Recent video LLMs introduce textual timestamps, which improve temporal grounding but further strain the token budget. 
Existing pruning retains selected visual tokens from each frame, discarding evidence while leaving timestamps underused as memory.
We instead view compression as evidence consolidation: discarded visual information should survive in retained visual tokens and timestamps, without training or additional long-term KV-cache positions. 
To make retained visual tokens effective evidence carriers, we propose \textbf{\pruner}, a causal pruner that protects temporal changes, selects representatives spanning diverse visual content, and folds similar discarded features into them.
To preserve fine details folding may obscure, we further introduce Timestamp Evidence Consolidation (\textbf{\method}), a plug-and-play module that exploits the frozen LLM's existing timestamp-attention pathway to transfer original dropped-token evidence into timestamps before eviction.
Together with \pruner, it improves Charades-STA mIoU by 9.03 over NovaCov at 12.5\% retention, reaching 94.7\% of unpruned accuracy, with 6.7\% extra per-step latency at one hour.
Our full approach achieves 4.6$\times$ faster per-frame processing than unpruned inference at 10 minutes and supports hour-long streams with only 14.44 GiB of GPU memory.
Code will be released at \url{https://github.com/JiaqiLi404/TEC.git}.
\end{abstract}

\section{Introduction}
\label{sec:intro}
\begin{figure}[t]
\centering
\includegraphics[width=0.95\columnwidth]{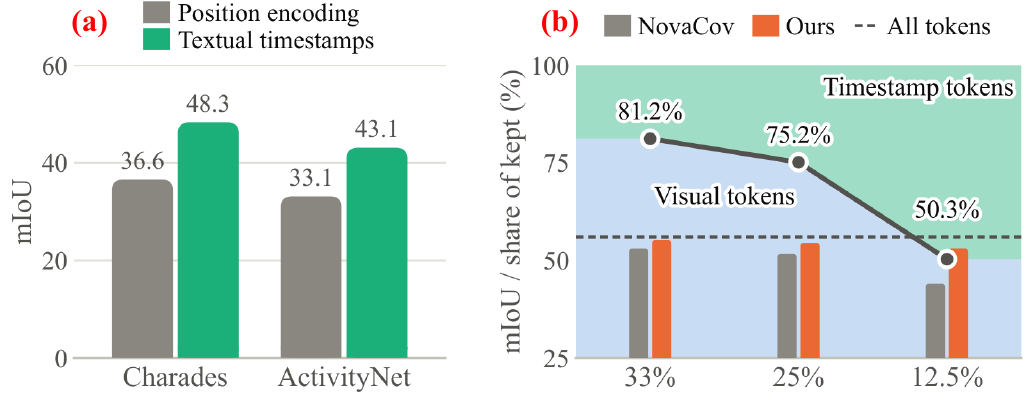}
\caption{\textbf{Textual timestamps strengthen temporal grounding but tighten the visual-token budget.} (a)~Performance of Qwen2.5-VL-7B with positional time encoding versus textual timestamps on TimeLens~\cite{zhang2026timelens}. (b)~The line shows the proportion of visual tokens among retained tokens at different retention ratios; the bars compare the mIoU of NovaCov \cite{duan2026novacov} and our method.}
\label{fig:intro-charts}
\end{figure}

Streaming video assistants enable real-time temporal understanding in robotics \cite{sermanet2024robovqa}, smart glasses \cite{yang2025egolife}, and live monitoring \cite{qian2025dispider}, answering questions about what happened and when as video unfolds. 
Streaming video LLMs must continuously process incoming frames while preserving past visual evidence \cite{chen2024videollm,huang2025online}. 
To this end, they accumulate a historical key--value (KV) cache that every subsequent frame must attend to and future questions rely on.
However, the growing history increases both memory consumption and per-frame attention cost. 
After one hour, a 4B model requires 57.2 GiB of memory and 4.4 seconds to process each new frame, making real-time streaming impractical~\cite{kim2026infinipot}.
Low-bit KV-cache quantization reduces memory usage but leaves attention costs largely unchanged and may introduce dequantization overhead \cite{hooper2024kvquant}.
To curb this growth, token compression prunes or merges redundant visual tokens, reducing LLM processing and KV-cache consumption~\cite{wang2026accelerating,chen2026streamingtom,duan2026novacov}. 
Yet aggressive compression risks discarding evidence crucial to temporal reasoning, raising the challenge of preserving task-relevant visual information under tight memory constraints~\cite{du2026unified}.

Time makes this harder. 
Recent video LLMs introduce textual timestamps to explicitly align visual observations with video time~\cite{qwen35blog,bai2025qwen3,zhang2026timelens}, achieving substantially better temporal grounding than implicit positional encoding, as shown in \cref{fig:intro-charts}(a). 
Yet these timestamps and their markers also consume scarce KV-cache positions, directly competing with visual tokens. 
As shown in \cref{fig:intro-charts}(b), they occupy almost half of the retained token budget at 12.5\% retention, leaving limited capacity for visual evidence and contributing to substantial performance degradation.

\begin{figure}[t]
\centering
\includegraphics[width=0.9\columnwidth]{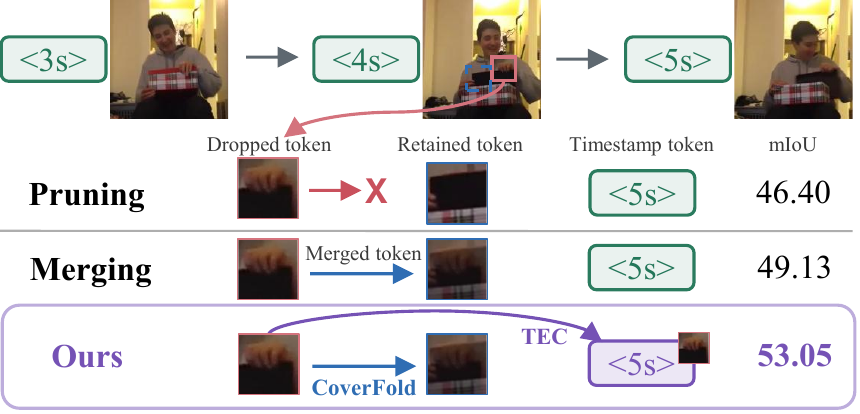}
\caption{\textbf{Keep the evidence, drop the tokens.} Pruning removes the highlighted visual token, potentially losing fine-grained evidence of small objects, while merging folds it into retained tokens, potentially blurring distinctive details. Our approach combines CoverFold's folding with TEC, which additionally transfers the original dropped evidence into timestamps before eviction.  }
\label{fig:teaser}
\end{figure}

However, existing video compression methods primarily optimize which visual tokens to retain or merge, largely leaving timestamps outside the compression objective~\cite{wang2026accelerating,chen2026streamingtom}. 
As illustrated in \cref{fig:teaser}, pruning discards small but potentially informative visual regions~\cite{wang2026accelerating}, while merging can obscure distinctive details through feature aggregation~\cite{bolya2022token}.
Such losses are particularly concerning for small objects, whose evidence may occupy only a single visual token. 
More fundamentally, both methods eliminate the discarded tokens' original attention pathways, preventing their evidence from directly enriching retained representations.
This reveals a critical missed opportunity that timestamps already consume substantial KV-cache positions, yet their capacity to absorb visual evidence remains untapped.

To address this gap, we rethink compression as not only deciding which tokens to retain, but also where discarded evidence can go without additional KV positions. 
We propose \pruner, a causal visual-token compressor that preserves temporal changes and selects representatives to fold similar discarded features. 
Yet merging may blur distinctive details. 
We therefore introduce Timestamp Evidence Consolidation (\method), a training-free, plug-and-play module that temporarily processes recent dropped tokens in early LLM layers, allowing retained tokens, especially timestamps, to absorb their original evidence before eviction. 

As illustrated in \cref{fig:teaser}, CoverFold and TEC mitigate the evidence loss of conventional pruning and merging~\cite{wang2026accelerating,bolya2022token} through visual folding and the LLM's existing attention pathways, helping preserve fine-grained evidence from small objects that might otherwise be lost.
Compared with unpruned inference, our approach retains only a fraction of the historical KV cache, while TEC recovers much of the lost accuracy with modest additional streaming latency by processing only recent dropped tokens.
At 12.5\% retention on Qwen3-VL-4B, our method gains 9.03 mIoU over NovaCov~\cite{duan2026novacov} on Charades-STA, retaining 94.7\% of unpruned accuracy.
It is 4.6$\times$ faster than unpruned inference at 10 minutes; at one hour, it uses 14.44\,GiB versus 57.17\,GiB while adding 6.7\% per-step latency over \pruner alone.

Our contributions are:
\begin{itemize}
\item We identify native timestamps as underused carriers of visual evidence, shifting the focus of streaming compression from which tokens to retain to how discarded evidence can survive within existing KV positions.
\item We realize this insight through \pruner, which preserves temporal changes and folds discarded features into visual tokens, and \method, which restores attention to dropped evidence through retained tokens, especially timestamps, without training or extra long-term KV memory.
\item At 12.5\% retention, we retain 94.7\% of unpruned performance with modest overhead, enabling hour-long streaming VLM inference on resource-constrained hardware.

\end{itemize}

\section{Related Work}
\label{sec:related}

\textbf{Visual Pruning.}
Visual-token pruning reduces LLM computation by retaining informative tokens based on attention-based importance~\cite{chen2024fastv,zhang2024sparsevlm,xing2024pyramiddrop} or visual coverage~\cite{zhang2025beyond,alvar2025divprune,wen2025stop,cho2026floc}.
While importance-based methods rank tokens individually, coverage-based methods select a set that collectively represents diverse visual content.
CoverPruner~\cite{zhu2026speaks} further formalizes this objective through query-weighted coverage.
However, these methods retain the chosen features without incorporating information from discarded tokens.

Streaming video poses a harder challenge, as future questions are unknown and compression must causally preserve both persistent scene context and transient changes.
To address these temporal requirements, DTD~\cite{yao2025timechat} prioritizes frame-to-frame changes, STC-Pruner~\cite{wang2026accelerating} measures novelty against spatial and temporal anchors, and NovaCov~\cite{duan2026novacov} combines current-frame coverage with historical novelty.
Despite their different selection criteria, these methods share a fundamental limitation that unselected visual tokens are discarded before reaching the LLM, potentially losing evidence crucial to future temporal reasoning. 
This motivates moving beyond token selection to preserving information from discarded features within the same token budget.

\textbf{Visual Merging.}
Token merging consolidates redundant visual features into retained tokens, as in ToMe~\cite{bolya2022token}, while PruMerge~\cite{shang2025llava} and VisionZip~\cite{yang2025visionzip} further combine pruning with merging.
ERA~\cite{wang2026era} and VPRune~\cite{lv2026vprune} further recover discarded evidence through attention rectification or feature recycling, but focus on retained visual representations rather than timestamp-based consolidation.
Video-oriented methods exploit spatial and temporal redundancy~\cite{tao2025dycoke,fu2025framefusion,huang2025prunevid,shao2026holitom}, but often rely on future frames or complete video clips, preventing causal streaming compression as frames arrive.

Moreover, aggregation may blur distinctive temporal details.
StreamingTOM~\cite{chen2026streamingtom} clusters static features and selects salient dynamic tokens, while CoViST~\cite{zhang2026covist} combines coverage-based selection with folding. 
Both may lose subtle temporal evidence during compression.
In contrast, \pruner protects temporal changes and selects coverage-maximizing representatives that absorb similar discarded features, whether static or dynamic.
This couples token selection with evidence consolidation under tight streaming budgets.
Yet folding removes the discarded tokens' original attention pathways, which \method temporarily restores.

\textbf{Evidence-Preserving Streaming Compression.}
Streaming compression must preserve visual evidence for unknown future questions while limiting the KV positions retained from each incoming frame.
InfiniPot-V~\cite{kim2026infinipot} bounds streaming memory through continual KV compression, but processes all visual tokens through the full LLM before eviction, retaining their computation cost.
SepLLM~\cite{chen2024sepllm} also supports streaming by consolidating text segments into native separators, but does not address visual evidence discarded by video token pruning.
Other methods exploit information transfer within the LLM: VTW~\cite{lin2025boosting} and ADSC~\cite{deniz2025vision} remove visual tokens after partial processing, while SIEVE~\cite{ou2026vision} uses visual information already absorbed by text to guide selection.
Learned visual carriers offer another means of consolidation~\cite{ye2025voco,zheng2026visco,wang2025seeing}.
However, these approaches do not specifically address how to recover the discarded evidence without changing its retained-token budget.

Our \method preserves the pruner's selections while temporarily processing its dropped tokens, allowing retained representations and timestamps to attend to their original evidence across recent frames.
Their temporary KV is subsequently evicted, preserving the long-term KV budget without retraining or full-depth processing of dropped tokens.

\textbf{Timestamps in Video LLMs.}
Temporal grounding increasingly relies on explicit textual timestamps and markers rather than positional encoding alone~\cite{guo2025vtg,huang2024lita,zeng2025timesuite,bai2025qwen3}.
TimeMarker~\cite{chen2024timemarker} combines temporal separators with adaptive visual compression.
More closely, Video-XL-2~\cite{qin2025video} uses historical timestamp KV states to convey coarse context across video chunks, reducing cross-chunk attention.
This demonstrates the potential of timestamps as contextual memory, but does not directly preserve the original visual evidence discarded by pre-LLM pruning.
Our \method instead exploits native timestamps as receivers of such dropped evidence during query-agnostic streaming, enriching existing KV positions without expanding the long-term cache.
\section{CoverFold}
\label{sec:method}

\begin{figure*}[ht]
  \centering  \includegraphics[width=0.98\textwidth]{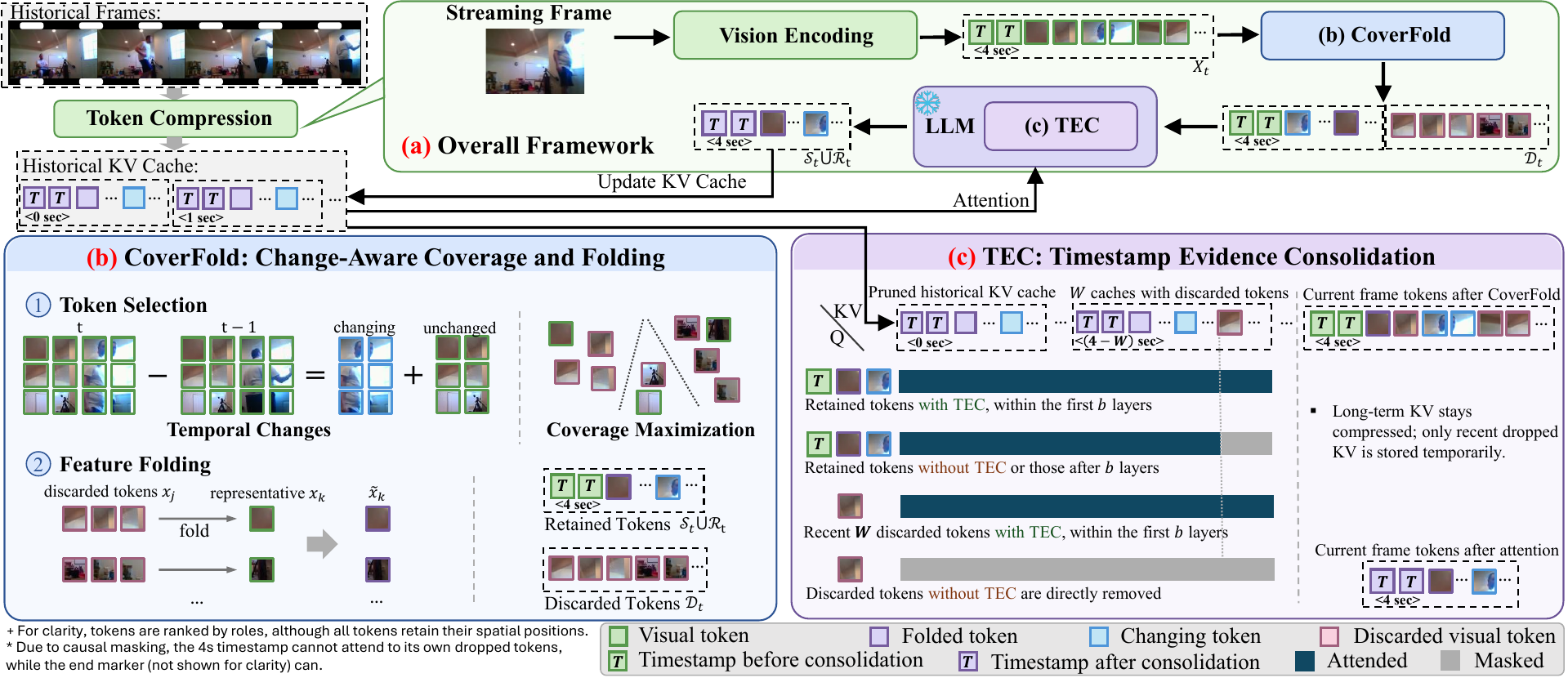}
  \caption{\textbf{Overview.} (a) In streaming LLM, each incoming frame attends to historical KV to build its cache. Maintaining a compact KV cache is thus essential. CoverFold compresses frames before LLM, where TEC enriches retained representations. (b) \textbf{CoverFold} prioritizes temporal changes, selects representatives maximizing visual coverage, and folds discarded features into them. (c) \textbf{TEC} allows retained tokens, particularly timestamps, to attend to recently discarded evidence in early LLM layers before evicting their temporary KV.}
  \label{fig:overview}
\end{figure*}

\subsection{Problem Definition}
\label{sec:setup}

We consider streaming video compression, where frames arrive at a fixed rate $f$.
A frozen visual encoder converts each incoming frame into $N_t=h\times w$ visual tokens $X_t$, where $h$ and $w$ denote the grid dimensions.
Each frame is accompanied by a textual timestamp (\eg, \texttt{<4.3 seconds>}), a start marker, and an end marker, collectively denoted as $\Sset_t$ with $|\Sset_t|=m_t$.
Given a retention rate $r$, the compressor preserves all timestamp tokens $\Sset_t$ and selects $q_t= r(N_t+m_t)-m_t$ visual tokens $\Rset_t\subseteq X_t$, discarding the rest $\Dset_t=X_t\setminus\Rset_t$.
The retained tokens are processed by the frozen LLM and stored in its layer-wise Key--Value (KV) cache.
When a question arrives, the model retrieves from this memory to generate an answer.
Importantly, compression must be causal, without access to future frames or queries.

However, each incoming frame must attend to previously cached tokens, causing per-frame processing latency to increase with video duration.
Tightening the retention budget helps meet the real-time deadline $1/f$, but risks losing evidence needed for future questions. 
We therefore seek to preserve information from discarded tokens $\Dset_t$ within retained representations, without storing their KV entries.

\textbf{Overview.}
As illustrated in \cref{fig:overview}, we use two complementary paths to preserve discarded evidence. 
Visual folding consolidates similar features, but requires representatives that cover diverse content without missing temporal changes.
This motivates \pruner's change-aware coverage objective. 
Since folding may obscure distinctive details, \method provides a second path through the frozen LLM's existing attention mechanism.
It lets retained tokens, particularly temporally anchored timestamps, to absorb evidence before the dropped KV is released. 
Both paths enrich retained representations within the same long-term memory budget.

\subsection{Change-Aware Coverage and Folding}
\label{sec:coverfold}

Most existing pruning methods discard unselected tokens~\cite{chen2024fastv,alvar2025divprune,wang2026accelerating}, while similarity-based merging may blur transient changes~\cite{bolya2022token,yang2025visionzip,chen2026streamingtom}, losing critical cues for temporal event understanding and localization.
To address both limitations, \pruner prioritizes changing regions and maximizes visual coverage when selecting representative tokens, before folding discarded tokens into their closest representatives.

\textbf{Temporal Changes.}
Changes provide important cues for event localization. 
Following SemVID~\cite{li2026keeping}, we quantify changes using the $\ell_2$ distance $d_{t,p}$ in consecutive frames:
\begin{equation}
d_{t,p}=\|x_{t,p}-x_{t-1,p}\|_2,
\label{eq:change}
\end{equation}
where $x_{t,p}$ denotes the feature at patch $p$ in frame $t$. 
We reserve $q_c=\gamma q_t$ tokens for the largest changes, where $\gamma\in[0,1]$ controls the fraction allocated to protecting transient information from subsequent merging. 
For the first frame, no change tokens are selected as no previous frame is available.

\textbf{Coverage Maximization.}
Token folding is most effective when discarded tokens have similar retained representatives; otherwise, their evidence may be distorted by merging.
We therefore use the remaining budget to maximize visual coverage $f(\Rset_t)$, measured by the weighted similarity of each token to its closest retained representative:
\begin{equation}
f(\Rset_t)=\sum_{i\in X_t}\omega_i\max_{j\in\Rset_t}s_{ij},\qquad\omega_i=(\|x_i\|/\bar n_t)^{\kappa},
\label{eq:cover}
\end{equation}
where $s_{ij}=[\cos(x_i,x_j)]_+$ measures feature similarity, and $\omega_i$ weights each token by its feature norm relative to the frame mean $\bar n_t$, with exponent $\kappa$. 
Negative similarities are clipped to zero, as dissimilar tokens are unsuitable for folding, meanwhile the weights prioritize salient content, using feature magnitude as a proxy for importance.

Starting from the change-selected tokens, we greedily add the token with the largest marginal gain until $|\Rset_t|=q_t$.
By aligning selection with the similarity used for folding, \pruner reduces redundant selections and provides better representatives for preserving discarded evidence.

\textbf{Feature Folding.}
With representatives selected to maximize coverage, we fold discarded tokens into their closest retained representatives using the feature similarity.
Specifically, each discarded token $j$ is assigned to its most similar coverage-selected token $k$ if their cosine similarity exceeds a threshold $\tau$; otherwise, it is left unmerged.

Let $\mathcal A_k$ denote the tokens assigned to $k$ and weight $w_{jk}=\cos(x_j,x_k)$. Their features are consolidated as
\begin{equation}
\bar x_k=\frac{x_k+\sum_{j\in\mathcal A_k}w_{jk}x_j}
{1+\sum_{j\in\mathcal A_k}w_{jk}},
\qquad
\widetilde x_k=\rho_k\frac{\bar x_k}{\|\bar x_k\|},
\label{eq:merge}
\end{equation}
where $\rho_k$ is the weighted mean feature norm, preventing magnitude shrinkage during averaging.

Folding merges tokens only within the same frame, replacing each retained token's features while preserving its original position. 
Thus, \pruner consolidates visual evidence without additional LLM computation or KV-cache entries. 
The original dropped features are separately passed to \method for further evidence consolidation. 

\subsection{Timestamp Evidence Consolidation}
\label{sec:tec}
While \pruner preserves discarded evidence through folding, averaging may obscure distinctive details. 
Pruning also prevents retained tokens, particularly timestamps, from attending to discarded visual tokens, disrupting an information pathway already used by the frozen LLM. 
Since timestamps are temporally anchored yet consume much of the token budget, enriching them with discarded visual evidence makes better use of scarce KV positions.

To illustrate this information loss, consider the attention output $o_i^{\ell}$ of a retained token at position $i$ (including timestamps) in layer $\ell$ of the unpruned LLM, which attends to preceding tokens through the historical KV cache:
\begin{equation}
o_i^{\ell}
= \sum_{j\in\Pset(i)}\alpha_{ij}^{\ell}v_j^{\ell}
+ \sum_{j\in\Dset(i)}\alpha_{ij}^{\ell}v_j^{\ell},
\label{eq:write}
\end{equation}
where $\Pset(i)$ and $\Dset(i)$ denote all preceding retained and dropped positions across frames, and $\alpha_{ij}^{\ell}$ and $v_j^{\ell}$ are attention weights and value vectors. 
Standard pruning discards visual tokens to reduce the KV cache, eliminating the second term and preventing their evidence from entering retained states.
\method lets tokens transfer evidence before eviction, preserving information without retaining their KV entries.

\textbf{Windowed Timestamp Attention.}
To recover discarded evidence without full-token processing, \method temporarily restores the dropped visual tokens from the most recent $W$ frames alongside the compressed historical KV cache.
Within the first $b$ LLM layers, retained tokens, particularly timestamps, attend to these original features, while dropped tokens attend to retained history and one another.
Dropped tokens are then evicted, whereas retained tokens continue through all $L$ layers, preserving the consolidated evidence without expanding permanent memory.
This approximates the unpruned attention in \cref{eq:write} through
\begin{equation}
\begin{gathered}
\tilde o_i^{\ell}=\sum_{j\in\Pset(i)}\tilde\alpha_{ij}^{\ell}v_j^{\ell}+\sum_{j\in\Dset(i)}g_{ij}^{\ell}\,\tilde\alpha_{ij}^{\ell}v_j^{\ell},\\
g_{ij}^{\ell}=\mathbf{1}[1\le\ell\le b]\;\mathbf{1}[t(i)-t(j)< W],
\end{gathered}
\label{eq:read}
\end{equation}
where $t(\cdot)$ denotes the frame index and $\tilde\alpha_{ij}^{\ell}$ is normalized over visible keys.



Compared with unpruned inference, \method maintains a compact historical KV cache containing only retained tokens, substantially reducing memory consumption and attention costs as the video grows. 
Compared with standard pruning, it temporarily restores access to dropped tokens from only recent $W$ frames in the first $b$ layers. 
This bounded additional computation and caching preserves substantially more visual evidence without restoring the full historical KV cache.

\textbf{Temporary Evidence Cache.}
To support attention over recent dropped tokens, \method temporarily retains their KV states alongside the compressed historical cache.
After frame $t$, with $\mathrm{KV}^{1:\ell}$ the keys and values of layers $1$ to $\ell$, the long-term memory $\mathcal M_t$ and the temporary cache $\mathcal C_t$ are
\begin{equation}
\begin{gathered}
\mathcal M_t=\mathcal M_{t-1}\cup\mathrm{KV}^{1:L}(\Sset_t\cup\Rset_t),\\
\mathcal C_t={\textstyle\bigcup_{s=t-W+1}^{t}}\,\mathrm{KV}^{1:b}(\Dset_s).
\end{gathered}
\label{eq:cache}
\end{equation}
During attention, both caches are accessed under the original token positions and causal mask, allowing timestamps to read recent dropped evidence as part of the historical context. 
Expired dropped states are evicted, and the temporary cache is cleared before question answering. 
Their consolidated evidence remains in the retained timestamp states, preserving the long-term memory size of standard pruning.

\method is training-free and compatible with other causal pruners.
The complete causal attention mask and procedure are detailed in the supplementary material.

\section{Experiments}
\label{sec:experiments}
\begin{table*}[t]
\caption{\textbf{Main results.}
(a) \textbf{Temporal grounding (mIoU) and question answering (accuracy) performance.}
Shaded rows add \method to the corresponding pruner. Within each model, \textbf{bold} and \underline{underlining} mark the \textbf{best} and \underline{second-best} compressed results.
(b) \textbf{Small-object grounding} on 854 Charades-STA queries: mIoU and its absolute drop.
(c) \textbf{Streaming efficiency} at 12.5\% retention: mean step latency and final-step memory.
Latency covers each method's complete compression and LLM execution, excluding vision encoding.
}
\label{tab:main}
\centering\footnotesize
\setlength{\tabcolsep}{3pt}
\renewcommand{\arraystretch}{0.92}
\newcommand{\tec}{\rowcolor{gray!13}\cellcolor{white}&\quad + \textbf{\method (ours)}}
\newcommand{\teclast}[1]{\rowcolor{gray!13}\cellcolor{white}#1&\quad + \textbf{\method (ours)}}
\resizebox{0.98\linewidth}{!}{%
\begin{tabular}{ll*{4}{>{\hspace{2pt}}w{c}{25pt}w{c}{25pt}<{\kern2pt}}|*{2}{>{\hspace{2pt}}w{c}{25pt}w{c}{25pt}<{\kern2pt}}|cccc}
\toprule
& & \multicolumn{8}{c}{\textbf{(a) Grounding and Question Answering}} & \multicolumn{4}{c}{\textbf{(b) Small Objects}} & \multicolumn{4}{c}{\textbf{(c) Efficiency}}\\
\cmidrule(lr){3-10}\cmidrule(lr){11-14}\cmidrule(lr){15-18}
& & \multicolumn{2}{c}{Charades-STA} & \multicolumn{2}{c}{ActivityNet} & \multicolumn{2}{c}{Video-MME} & \multicolumn{2}{c}{LongVB} & \multicolumn{2}{c}{12.5\%} & \multicolumn{2}{c}{25\%} & \multicolumn{2}{c}{10 min} & \multicolumn{2}{c}{1 hour}\\
\cmidrule(lr){3-4}\cmidrule(lr){5-6}\cmidrule(lr){7-8}\cmidrule(lr){9-10}\cmidrule(lr){11-12}\cmidrule(lr){13-14}\cmidrule(lr){15-16}\cmidrule(lr){17-18}
Model & Method & 12.5\% & 25\% & 12.5\% & 25\% & 12.5\% & 25\% & 12.5\% & 25\% & mIoU & $\Delta\downarrow$ & mIoU & $\Delta\downarrow$ & ms & GiB & ms & GiB\\
\midrule
& All tokens & \multicolumn{2}{c}{55.99} & \multicolumn{2}{c}{42.81} & \multicolumn{2}{c}{66.81} & \multicolumn{2}{c}{61.11} & \multicolumn{4}{c}{53.14} & 206.9 & 16.35 & 4380.2$^\S$ & 57.17$^\S$\\
 & HoliTom~\cite{shao2026holitom}$^\dagger$ & 45.93 & 52.86 & 37.48 & 41.10 & 57.22 & 63.63 & 55.72 & 58.71 & 43.36 & 9.78 & 50.18 & 2.96 & -- & -- & -- & --\\
 & FlashVID~\cite{fan2026flashvid}$^\dagger$ & 46.11 & 53.59 & 37.83 & 41.88 & 58.74 & 64.52 & 56.47 & 59.54 & 43.71 & 9.43 & 50.94 & 2.20 & -- & -- & -- & --\\
 & SemVID~\cite{li2026keeping}$^\dagger$ & 48.41 & 53.54 & 38.96 & 41.90 & 56.78 & 62.81 & 55.35 & 59.99 & 46.95 & 6.19 & 50.70 & 2.44 & -- & -- & -- & --\\
 & InfiniPot-V~\cite{kim2026infinipot} & 50.89 & 54.33 & 39.11 & 41.81 & 59.85 & 65.23 & 55.65 & 58.64 & 49.05 & 4.09 & 51.22 & 1.92 & 141.4 & 9.43 & 1122.7 & 15.12\\
 & HERMES~\cite{zhang2026hermes} & 51.06 & 53.81 & 39.65 & 42.13 & 59.33 & 63.33 & 52.73 & 54.75 & 47.48 & 5.66 & 50.58 & 2.56 & 111.3 & 9.30 & 317.3 & 14.45\\
 & FastVID~\cite{shen2026fastvid} & 37.06 & 52.07 & 36.20 & 41.48 & 55.44 & 60.63 & 53.25 & 58.79 & 34.63 & 18.51 & 49.02 & 4.12 & 40.8 & 9.29 & 154.0 & 14.40\\
 & DivPrune~\cite{alvar2025divprune} & 40.91 & 50.73 & 36.19 & 41.59 & 59.41 & 64.41 & 53.85 & 58.56 & 37.65 & 15.49 & 46.84 & 6.30 & 39.7 & 9.29 & 153.3 & 14.40\\
 & STC-Pruner~\cite{wang2026accelerating} & 35.58 & 49.85 & 31.65 & 40.41 & 59.41 & 63.89 & 51.53 & 57.97 & 35.99 & 17.15 & 48.05 & 5.09 & 40.1 & 9.29 & 153.8 & 14.40\\
\tec & 51.41 & 53.31 & 38.28 & 41.69 & 60.59 & 64.04 & 52.88 & 57.82 & 49.56 & 3.58 & 50.90 & 2.24 & 44.5 & 9.32 & 164.1 & 14.44\\
 & VisionZip~\cite{yang2025visionzip} & 37.52 & 51.56 & 36.14 & 40.65 & 53.78 & 62.19 & 52.28 & 56.10 & 36.32 & 16.82 & 49.11 & 4.03 & 39.6 & 9.29 & 153.3 & 14.40\\
\tec & 52.16 & 54.26 & 39.82 & 42.33 & 55.74 & 62.81 & 52.80 & 56.92 & 49.06 & 4.08 & 51.62 & 1.52 & 43.9 & 9.32 & 163.6 & 14.44\\
 & StreamingTOM~\cite{chen2026streamingtom} & 38.93 & 51.27 & 36.56 & 40.87 & 54.37 & 62.59 & 52.73 & 56.92 & 38.19 & 14.95 & 48.58 & 4.56 & 39.9 & 9.29 & 153.5 & 14.40\\
\tec & 52.23 & 54.03 & 40.51 & 41.81 & 56.37 & 63.19 & 53.40 & 57.67 & 49.48 & 3.66 & 51.45 & 1.69 & 44.3 & 9.32 & 163.8 & 14.44\\
 & DTD~\cite{yao2025timechat} & 39.07 & 50.35 & 34.82 & 39.78 & 58.52 & 63.48 & 52.95 & 58.04 & 34.94 & 18.20 & 47.55 & 5.59 & 39.5 & 9.29 & 153.2 & 14.40\\
\tec & 52.31 & 54.08 & 40.34 & 42.08 & 60.85 & 64.07 & 53.85 & 57.44 & \underline{50.19} & \underline{2.95} & \underline{51.75} & \underline{1.39} & 43.8 & 9.32 & 163.5 & 14.44\\
 & ToMe~\cite{bolya2022token} & 42.46 & 52.04 & 40.48 & \underline{43.06} & 58.26 & 63.59 & 55.65 & 59.54 & 36.64 & 16.50 & 48.31 & 4.83 & 42.6 & 9.29 & 156.3 & 14.40\\
\tec & 52.32 & 54.00 & 41.34 & 42.92 & 58.96 & 63.93 & 56.54 & \underline{60.13} & 48.49 & 4.65 & 50.22 & 2.92 & 47.0 & 9.32 & 166.6 & 14.44\\
 & NovaCov~\cite{duan2026novacov} & 44.02 & 51.65 & 39.13 & 41.79 & 60.63 & 65.22 & 56.39 & 58.86 & 38.99 & 14.15 & 47.89 & 5.25 & 41.1 & 9.29 & 154.4 & 14.40\\
\tec & \underline{52.98} & \textbf{54.50} & \underline{41.42} & 42.86 & \underline{61.67} & \underline{65.56} & 55.65 & 59.16 & 49.36 & 3.78 & 51.04 & 2.10 & 45.5 & 9.32 & 164.7 & 14.44\\
 & \textbf{\pruner (ours)} & 49.13 & 53.57 & 40.76 & 42.84 & 60.97 & 65.22 & \underline{57.22} & 59.84 & 45.55 & 7.59 & 49.93 & 3.21 & 41.0 & 9.29 & 154.3 & 14.40\\
\teclast{\multirow{-21}{*}{\shortstack{Qwen3\\4B}}} & \textbf{53.05} & \underline{54.45} & \textbf{42.01} & \textbf{43.53} & \textbf{61.82} & \textbf{65.86} & \textbf{57.97} & \textbf{60.73} & \textbf{50.25} & \textbf{2.89} & \textbf{51.89} & \textbf{1.25} & 45.3 & 9.32 & 164.6 & 14.44\\

\midrule
& All tokens & \multicolumn{2}{c}{56.67} & \multicolumn{2}{c}{44.03} & \multicolumn{2}{c}{68.30} & \multicolumn{2}{c}{62.68} & \multicolumn{4}{c}{54.27} & 203.4 & 24.42 & 4370.3$^\S$ & 65.24$^\S$\\
 & VisionZip & 36.71 & 53.44 & 33.91 & 42.53 & 55.96 & 65.15 & 53.25 & 57.44 & 33.09 & 21.18 & 52.21 & 2.06 & 43.1 & 17.35 & 157.9 & 22.46\\
\tec & 53.75 & 55.69 & 39.88 & 43.51 & 59.15 & 65.48 & 52.36 & 58.64 & 51.09 & 3.18 & \underline{53.26} & \underline{1.01} & 48.9 & 17.38 & 170.2 & 22.50\\
 & DTD & 35.76 & 50.63 & 32.74 & 40.22 & 58.89 & 64.15 & 52.51 & 57.59 & 30.29 & 23.98 & 47.51 & 6.76 & 43.0 & 17.35 & 157.8 & 22.46\\
\tec & 53.95 & 55.37 & 40.56 & 42.15 & 61.67 & 65.41 & 53.03 & 58.12 & 50.96 & 3.31 & 52.76 & 1.51 & 48.8 & 17.38 & 170.2 & 22.50\\
 & ToMe & 43.35 & 52.30 & 38.37 & 42.89 & 60.48 & 65.56 & 55.65 & 58.79 & 36.44 & 17.83 & 48.97 & 5.30 & 46.0 & 17.35 & 161.3 & 22.46\\
\tec & 54.11 & 55.55 & 41.03 & 43.45 & 60.78 & 64.89 & 56.17 & 59.09 & 50.72 & 3.55 & 52.87 & 1.40 & 51.9 & 17.38 & 173.7 & 22.50\\
 & NovaCov & 44.68 & 51.68 & 38.30 & 42.32 & 61.15 & 65.30 & 56.17 & 60.59 & 38.96 & 15.31 & 48.39 & 5.88 & 44.5 & 17.35 & 159.0 & 22.46\\
\tec & \underline{54.60} & \underline{55.88} & \textbf{41.97} & 43.36 & \textbf{62.56} & \underline{65.63} & 55.57 & 60.06 & \underline{51.72} & \underline{2.55} & 53.21 & 1.06 & 50.4 & 17.38 & 171.3 & 22.50\\
 & \textbf{\pruner (ours)} & 48.41 & 54.16 & 40.45 & \underline{43.55} & 61.18 & 65.53 & \underline{57.52} & \underline{60.72} & 44.37 & 9.90 & 51.11 & 3.16 & 44.3 & 17.35 & 159.1 & 22.46\\
\teclast{\multirow{-11}{*}{\shortstack{Qwen3\\8B}}} & \textbf{54.64} & \textbf{55.90} & \underline{41.66} & \textbf{43.98} & \underline{62.39} & \textbf{65.70} & \textbf{58.12} & \textbf{60.73} & \textbf{52.29} & \textbf{1.98} & \textbf{53.33} & \textbf{0.94} & 50.2 & 17.38 & 171.5 & 22.50\\

\midrule
& All tokens & \multicolumn{2}{c}{61.25} & \multicolumn{2}{c}{46.34} & \multicolumn{2}{c}{69.63} & \multicolumn{2}{c}{62.60} & \multicolumn{4}{c}{58.70} & 73.8 & 19.37 & 407.1 & 28.50\\
 & VisionZip & 49.16 & 57.82 & 40.45 & 44.64 & 58.37 & 65.96 & 50.64 & 56.25 & 47.32 & 11.38 & 54.67 & 4.03 & 55.5 & 17.79 & 64.1 & 18.94\\
\tec & 58.35 & 59.36 & 42.82 & 44.82 & 60.22 & 66.37 & 51.23 & 56.17 & 55.05 & 3.65 & 56.39 & 2.31 & 57.6 & 17.81 & 76.9 & 18.97\\
 & DTD & 45.72 & 56.43 & 37.68 & 41.81 & 61.26 & 66.59 & 53.33 & 56.17 & 42.41 & 16.29 & 52.80 & 5.90 & 55.4 & 17.79 & 64.0 & 18.94\\
\tec & 57.83 & 59.56 & 42.83 & 44.44 & 62.11 & 66.26 & 52.51 & 54.90 & 54.18 & 4.52 & \textbf{56.42} & \textbf{2.28} & 57.5 & 17.81 & 76.8 & 18.97\\
 & ToMe & 50.39 & 58.01 & 41.92 & 44.44 & 61.81 & 66.33 & 54.82 & 57.82 & 46.74 & 11.96 & 54.90 & 3.80 & 58.3 & 17.79 & 66.9 & 18.94\\
\tec & 58.23 & \underline{59.61} & 43.12 & 44.89 & 61.85 & 66.37 & 54.60 & 58.49 & 54.89 & 3.81 & 56.34 & 2.36 & 60.4 & 17.81 & 79.7 & 18.97\\
 & NovaCov & 53.73 & 58.05 & 41.04 & 44.53 & 63.26 & 66.41 & 55.05 & 56.54 & 49.80 & 8.90 & 54.09 & 4.61 & 56.8 & 17.79 & 65.1 & 18.94\\
\tec & \underline{58.46} & 59.59 & \underline{43.37} & 44.71 & \underline{63.63} & \textbf{67.33} & 55.08 & 56.61 & \underline{55.14} & \underline{3.56} & 56.37 & 2.33 & 58.9 & 17.81 & 77.9 & 18.97\\
 & \textbf{\pruner (ours)} & 55.01 & 58.54 & 42.06 & \textbf{45.50} & 63.42 & 66.52 & \underline{56.54} & \underline{58.61} & 50.68 & 8.02 & 55.20 & 3.50 & 56.7 & 17.79 & 65.1 & 18.94\\
\teclast{\multirow{-11}{*}{\shortstack{Qwen3.5\\9B}}} & \textbf{58.63} & \textbf{59.67} & \textbf{43.43} & \underline{45.09} & \textbf{63.92} & \underline{67.21} & \textbf{56.69} & \textbf{58.86} & \textbf{55.29} & \textbf{3.41} & \underline{56.40} & \underline{2.30} & 58.7 & 17.81 & 77.9 & 18.97\\
\bottomrule
\end{tabular}}
\par\vspace{2pt}
\parbox{\linewidth}{\scriptsize
$^\dagger$Non-streaming methods with future frames or query access.
$^\S$The uncompressed run requires host KV offloading and the reported memory includes GPU and host storage.
}

\end{table*}

\begin{table}[t]
\caption{\textbf{\method across retention budgets.} \#Visual denotes visual tokens per step, while the remainings are timestamps and anchors.}
\label{tab:budget-curve}
\centering\small
\setlength{\tabcolsep}{3.5pt}
\renewcommand{\arraystretch}{0.85}
\resizebox{0.86\linewidth}{!}{%
\begin{tabular}{lcccc}
\toprule
Charades-STA & \#Visual & \pruner & + \method & $\Delta$\\
\midrule
1 visual token per step ($\approx$ 6\%) & 1.0 & 24.23 & 50.58 & +26.35\\
12.5\% & 9.0 & 49.13 & 53.05 & +3.92\\
18.75\% & 17.8 & 52.27 & 54.08 & +1.81\\
25\% & 26.6 & \textbf{53.57} & \textbf{54.45} & +0.88\\
\midrule
All tokens & 132.3 & \multicolumn{2}{c}{55.99}\\
\bottomrule
\end{tabular}}
\end{table}

\subsection{Implementation Details}
\label{sec:exp-setup}

\textbf{Benchmarks.}
We evaluate temporal grounding on Charades-STA~\cite{gao2017charades} (3,720 queries, 1,334 indoor-activity videos) and ActivityNet Captions~\cite{krishna2017dense} (2,142 queries, 600 longer videos), using mean temporal intersection-over-union (mIoU) following~\citet{li2026keeping}.
We further assess small-object evidence preservation on a Charades-STA subset identified using Action Genome bounding-box annotations~\cite{ji2020action}.
For long-video Question Answering (QA), we evaluate on Video-MME~\cite{fu2025video} (2,700 questions across 900 diverse videos) and LongVideoBench~\cite{wu2024longvideobench} (1,337 questions across 753 videos, emphasizing fine-grained and cross-moment reasoning).
Both use multiple-choice accuracy without subtitles.
Further details are provided in the supplementary material.

\textbf{Models.}
Our primary backbone is frozen Qwen3-VL-4B~\cite{bai2025qwen3}, with Qwen3-VL-8B and Qwen3.5-9B~\cite{qwen35blog} testing generalization across model scales and architectures.

\textbf{Streaming Compression Protocol.}
We simulate streaming by processing videos incrementally, compressing the KV cache without access to future frames or questions.
Queries arrive after video encoding, evaluating the evidence preserved by causal, query-agnostic compression.
For temporal grounding, we sample at 2 fps (two frames per timestamp) following~\cite{yao2025timechat}.
For question answering, we follow ReKV~\cite{di2025streaming}, encoding videos once at 0.5 fps and retrieving historical steps when queried.
Both tasks use 12.5\% and 25\% retention budgets including timestamps and markers.
Further protocol details are in the supplementary material.

\textbf{Hyperparameters.}
We select parameters on held-out Charades-STA training videos for grounding and 500 publicly annotated EgoSchema questions~\cite{mangalam2023egoschema} for QA.
For grounding, \pruner uses a temporal-change fraction $\gamma=0.4$ (\cref{eq:change}), coverage exponent $\kappa=0.5$ (\cref{eq:cover}), and folding threshold $\tau=0$ (\cref{eq:merge}).
For QA, we set $\gamma=1$ and disable folding, prioritizing fine-grained evidence fidelity over the temporal coverage.
For Qwen3-VL-4B/8B, \method uses the first $b=18$ layers and the most recent $W=4$ steps (\cref{eq:read,eq:cache}).
For hybrid-attention Qwen3.5-9B, evidence is consolidated into recurrent states as detailed in the supplementary material.

\subsection{Main Results}
\label{sec:main}
Our approach preserves discarded visual evidence through two complementary components.
\pruner folds discarded features into selected visual representatives, while \method lets the original discarded features enter the frozen LLM before eviction to enrich retained states, particularly timestamps.

\textbf{TEC.}
TEC substantially improves existing pruners, especially under aggressive compression.
As shown in \cref{tab:main}(a), adding \method to STC-Pruner raises mIoU from 35.58 to 51.41 on Charades-STA and from 31.65 to 38.28 on ActivityNet at 12.5\% retention.
At the extreme budget of just one visual token per step in \cref{tab:budget-curve}, \method improves \pruner from 24.23 to 50.58, even surpassing nine visual tokens per step.
This indicates that although CoverFold folds discarded features into retained representatives, the original features still contain useful evidence that the LLM can absorb before eviction.

These results expose a key limitation of conventional pruning: as retention drops from 25\% to 12.5\%, NovaCov's mIoU falls from 51.65 to 44.02, as more evidence is discarded while timestamps occupy a larger share of retained positions.
TEC instead leverages these existing timestamps to absorb discarded visual evidence through the frozen LLM's attention, preserving information that conventional pruning would otherwise lose.
Conversely, at 25\% retention, where more visual evidence survives, TEC's gain over CoverFold drops from 3.93 to 0.88 mIoU but remains significant.

Additionally, TEC generalizes across pruning strategies, model scales, and tasks, improving most configurations on all four benchmarks.
More importantly, it reduces dependence on the initial token selection.
On Charades-STA at 12.5\% retention, the performance spread across seven pruners contracts from 13.55 (35.58--49.13) to just 1.64 (51.41--53.05).
This suggests that consolidating discarded evidence can compensate for imperfect selection.
Occasionally, TEC lowers accuracy, possibly because the pruner has already kept the key evidence and extra information distracts the model.
However, without knowing future questions, the pruner cannot tell which evidence is safe to discard.

\textbf{CoverFold.}
Complementing TEC's in-LLM consolidation, \pruner preserves evidence before LLM without additional LLM computation.
Unlike NovaCov, which selects representative tokens but discards the rest, or ToMe, which merges similar tokens without explicitly protecting temporal changes, \pruner combines both strengths.
It preserves changing regions while selecting coverage-maximizing representatives that can absorb similar discarded features.
This retains distinctive temporal cues while recovering evidence that conventional pruning would lose.
On Charades-STA at 12.5\% retention in \cref{tab:main}(a), \pruner achieves 49.13 mIoU, compared with 44.02 for NovaCov and 42.46 for ToMe.
Since folding can still blur fine details through averaging, \method further improves mIoU from 49.13 to 53.05 by incorporating the original discarded features.

\textbf{Small Objects.}
Small objects are particularly vulnerable to compression, as their evidence may occupy only a few visual tokens that pruning discards or folding dilutes.
As shown in \cref{tab:main}(b), for objects occupying less than 1\% of the frame, \method improves \pruner from 45.55 to 50.25 mIoU at 12.5\% retention, reducing the gap to uncompressed performance from 7.59 to 2.89.
This improvement holds across all tested pruners and backbones, highlighting the value of exposing original discarded features to the LLM rather than relying solely on selected or averaged representations.

\textbf{Question Answering.}
\pruner with \method achieves the highest compressed accuracy on LongVideoBench across all three backbones and both budgets according to \cref{tab:main}(a).
However, TEC brings smaller gains than in grounding.
On Qwen3-VL-4B at 12.5\% retention, accuracy improves from 57.22 to 57.97 (+0.75), compared with 49.13 to 53.05 (+3.92 mIoU) on Charades-STA.
This likely reflects different evidence requirements that temporal grounding relies on frame-to-frame changes to locate event boundaries, favoring broad temporal coverage, whereas QA often hinges on precise details at relevant moments, such as object attributes and spatial relations \cite{li2026keeping}.
This is consistent with our hyperparameter tuning, which favors disabling folding to preserve fidelity.

\textbf{Efficiency.}
As shown in \cref{tab:main}(c), our approach achieves 4.6$\times$ faster per-step processing than unpruned inference at 10 minutes on Qwen3-VL-4B at 12.5\% retention.
At one hour, it reduces memory usage from 57.17 to 14.44\,GiB.
The compressed history therefore fits on the GPU, avoiding host KV offloading and lowering latency from 4380.2 to 164.6\,ms per step and enabling efficient long-duration streaming.

Importantly, TEC adds only modest overhead over \pruner.
By limiting discarded-token processing to the first $b=18$ layers and retaining a recent window of $W=4$ steps, 
TEC preserves the compact long-term KV budget as other pruners, avoiding the dominant source of memory growth in streaming video.
Consequently, TEC adds only 4.3\,ms per step and 0.03\,GiB at 10 minutes, increasing to just 10.3\,ms (6.7\%) and 0.04\,GiB at one hour.
These results demonstrate TEC’s favorable balance between accuracy and efficiency.

\subsection{Ablations of TEC}
\label{sec:tec-analysis}
\label{sec:memory}
\begin{table}[t]
\caption{\textbf{\method ablations.}
In (b), VTW processes each step through $b$ layers, then discards all visual tokens.
In (c), ``digits 0'' replaces timestamp digits with zero.
In (d), blocking removes visual access from timestamp text and markers.
Costs are 20-minute mean ms/step and whole-hour LLM TFLOPs as detailed in Appx. E.}
\label{tab:tec-ablation}
\centering
\centering\footnotesize
\setlength{\tabcolsep}{2.5pt}
\resizebox{\linewidth}{!}{%
\begin{tabular}{@{}clcc|cc@{}}
\toprule
& Charades-STA, 12.5\% & mIoU & $\Delta$ [95\% CI]& ms/step & TFLOPs\\
\midrule
\multirow{5}{*}{\shortstack[c]{(a)\\\scriptsize Evidence\\\scriptsize and timing}} & Default \method & \textbf{53.05} & ref.&--&--\\
& Without folding & 52.57 & \ci{$-$0.48}{$-$1.05}{+0.09}&--&--\\
& Replace discarded features & 29.43 & \ci{$-$23.62}{$-$25.07}{$-$22.11}&--&--\\
& Read 4 steps late, flushed & 51.21 & \ci{$-$1.84}{$-$2.55}{$-$1.12}&--&--\\
& Read 8 steps late, flushed & 50.55 & \ci{$-$2.50}{$-$3.30}{$-$1.71}&--&--\\
\midrule
\multirow{4}{*}{\shortstack[c]{(b)\\\scriptsize Reader tokens}} & Default \method (all readers) & \textbf{53.05} & ref.&--&--\\
& Timestamp text + markers & 52.53 & \ci{$-$0.52}{$-$1.08}{+0.08}& 49.9 & 4,085\\
& Kept visual tokens only & 50.52 & \ci{$-$2.53}{$-$3.41}{$-$1.63}&--&--\\
& Current end marker & 48.40 & \ci{$-$4.65}{$-$5.50}{$-$3.76}&--&--\\
& VTW~\cite{lin2025boosting}, w/o pruner & 44.31 & \ci{$-$4.82}{$-$5.80}{$-$3.81} & 47.5 & 3,955\\
\midrule
\multirow{4}{*}{\shortstack[c]{(c)\\\scriptsize Timestamp\\\scriptsize digits}} & Default \method, native digits & 53.05 & --&--&--\\
& All tokens, digits 0 & 9.77 & --&--&--\\
& Pruning, digits 0 & 3.62 & ref.&--&--\\
& \method, digits 0 & \textbf{8.12} & \ci{+4.50}{+3.99}{+5.00}&--&--\\
\midrule
\multirow{3}{*}{\shortstack[c]{(d)\\\scriptsize Layer access}} & All tokens, no blocking & 55.99 & ref.&--&--\\
& Block layers 1--18 & 52.28 & \ci{$-$3.71}{$-$4.46}{$-$2.93}&--&--\\
& Block layers 19--36 & 56.05 & \ci{+0.06}{$-$0.35}{+0.48} &--&--\\
\midrule
\multirow{6}{*}{\shortstack[c]{(e)\\\scriptsize Layer depth\\\scriptsize\cref{eq:read}}}& $b=0$ (w/o \method) & 49.13 & ref. & 46.2 & 905\\
& $b=6$ & 49.43 & \ci{+0.30}{$-$0.14}{+0.73} & 48.4 & 1,965\\
& $b=12$ & 50.50 & \ci{+1.37}{+0.77}{+1.97} & 48.9 & 3,025\\
& $\boldsymbol{b=18}$ & 53.05 & \ci{+3.92}{+3.05}{+4.77} & 49.9 & 4,085\\
& $b=24$ & \textbf{53.48} & \ci{+4.35}{+3.44}{+5.23} & 51.3 & 5,146\\
& $b=36$ & 53.41 & \ci{+4.28}{+3.37}{+5.17} & 51.8 & 7,266\\
\midrule
\multirow{2}{*}{\shortstack[c]{(f)\\\scriptsize Beginning Layer}}& $\boldsymbol{a=1}$ & \textbf{53.05} & \ci{+3.92}{+3.05}{+4.77} & 49.9 & 4,085\\
& $a=13$ & 52.32 & \ci{+3.19}{+2.44}{+3.92} & 49.9 & 4,085\\
\midrule
\multirow{4}{*}{\shortstack[c]{(g)\\\scriptsize Reader window\\\scriptsize\cref{eq:read}}}& $W=0$ (w/o \method) & 49.13 & ref. & 46.2 & 905\\
& $W=2$ & 52.60 & \ci{+3.47}{+2.63}{+4.29} & 49.9 & 4,067\\
& $\boldsymbol{W=4}$ & \textbf{53.05} & \ci{+3.92}{+3.05}{+4.77} & 49.9 & 4,085\\
& $W=8$ & 52.27 & \ci{+3.14}{+2.29}{+4.00} & 50.0 & 4,121\\
\bottomrule
\end{tabular}}

\end{table}

\textbf{Evidence and timing.}
As shown in \cref{tab:tec-ablation}(a), removing folding causes only a small drop (53.05 to 52.57 mIoU), whereas replacing discarded features with those from half a video away sharply reduces performance to 29.43.
This confirms that TEC's gains depend on transferring meaningful discarded evidence, rather than merely adding tokens to attention.
Moreover, delaying access to the correct evidence by four or eight steps reduces mIoU to 51.21 and 50.55, respectively, even when all evidence is eventually processed.
Together, these results establish the importance of both accurate evidence and timely consolidation into retained states.

\textbf{Evidence Readers.}
As shown in \cref{tab:tec-ablation}(b), restricting consolidation to timestamp text and markers achieves 52.53 mIoU, close to 53.05 when all retained tokens participate.
In contrast, consolidation only to retained visual tokens reduces mIoU to 50.52.
This establishes timestamps as the primary carriers of discarded visual evidence.
Furthermore, replacing TEC-updated timestamp KV states with those from pruning reduces mIoU from 53.05 to 49.30 (supplementary material), confirming that timestamps retain the recovered evidence.

Despite comparable early-layer design, VTW~\cite{lin2025boosting} achieves only 44.31 mIoU.
This suggests that retained visual tokens still carry critical evidence after early-layers, which is lost to deeper layers when VTW withdraws them.

To distinguish the role of timestamps from their capacity to store visual evidence, we replace all timestamp digits with zero.
This reduces uncompressed mIoU from 55.99 to 9.77, yet TEC still improves zero-digit pruning from 3.62 to 8.12 (\cref{tab:tec-ablation}(c)).
These results show that while timestamps are essential for grounding, TEC can exploit their states to preserve discarded evidence even without valid time labels.

\textbf{Layers.}
In \cref{tab:tec-ablation}(d), we examine how timestamps absorb visual evidence by blocking their attention to visual tokens at different layers in the unpruned model.
Blocking layers 1--18 reduces mIoU from 55.99 to 52.28, whereas blocking layers 19--36 has no significant effect.
This indicates that timestamps primarily integrate evidence in early LLM layers.

Consistent with this finding, TEC achieves a 3.92 mIoU gain over pruning with $b=18$ according to \cref{tab:tec-ablation}(e), while extending processing to all 36 layers adds only 0.36 mIoU at 3.8\% higher latency.
Delaying access until layer 13 further reduces mIoU by 0.73 without saving computation, as discarded tokens must still pass through earlier layers (\cref{tab:tec-ablation}(f)).
Thus, TEC benefits from early evidence integration followed by eviction, avoiding unnecessary full-depth processing.

\textbf{Window.}
In \cref{tab:tec-ablation}(g), we vary how far back retained tokens can attend to discarded evidence.
Increasing $W$ from 2 to 4 improves mIoU from 52.60 to 53.05, while $W=8$ reduces it to 52.27.
A shorter window misses useful evidence, whereas a longer one may introduce stale or redundant information already consolidated into retained states.
Additionally, extending the window adds little latency, since discarded tokens are processed only once and $W$ mainly determines how long their temporary KV states are cached.

\textbf{TFLOPs.}
\label{sec:cost}
As detailed in Appx.~E, memory operations take $\approx88\%$ of TEC's step time, while computation takes only 5.2\,ms.
Consequently, despite 4.5$\times$ more FLOPs, TEC adds only 15.8\,ms per step, primarily due to cache copies.
Additional tokens share the same forward pass and model-weight accesses, while the permanent KV cache remains compact, limiting memory overhead.
Thus, FLOPs do not translate proportionally into latency.
Furthermore, adjusting the processing depth $b$ offers a flexible accuracy--latency trade-off for resource-constrained hardware (\cref{tab:tec-ablation}(e)).

\subsection{Ablations of CoverFold}
\label{sec:coverfold-analysis}
\label{sec:design}
\begin{table}[t]
\caption{\textbf{\pruner ablations.} (a) Component removal and (b) temporal-change fraction $\gamma$ (\cref{eq:change}).}
\label{tab:coverfold-ablation}
\centering
\centering\footnotesize
\setlength{\tabcolsep}{2.5pt}
\resizebox{0.96\linewidth}{!}{%
\begin{tabular}{@{}clccc@{}}
\toprule
&Charades-STA, 12.5\% & mIoU & $\Delta$ [95\% CI] & + \method\\
\midrule
\multirow{6}{*}{\shortstack[c]{(a)}} &\pruner & \textbf{49.13} & ref. & \textbf{53.05}\\
&\quad w/o folding ($\gamma=0.4$) & 46.40 & \ci{$-$2.73}{$-$3.54}{$-$1.92} & 52.57\\
&\quad w/o change tokens ($\gamma=0$) & 47.74 & \ci{$-$1.39}{$-$2.23}{$-$0.56} & 52.37\\
&\quad change tokens only ($\gamma=1$) & 48.93 & \ci{$-$0.20}{$-$0.41}{$-$0.02} & 52.64\\
&\quad w/o coverage, uniform selection & 48.53 & \ci{$-$0.60}{$-$1.34}{+0.12} & 52.86\\
&\quad unweighted coverage ($\kappa=0$) & 49.00 & \ci{$-$0.13}{$-$0.60}{+0.34} & 52.94\\
\midrule
\multirow{5}{*}{\shortstack[c]{(b)}} &$\gamma=0.2$ & 48.87 & \ci{$-$0.26}{$-$0.96}{+0.43} &52.75\\
&$\boldsymbol{\gamma=0.4}$ & \textbf{49.13} & ref.&\textbf{53.05}\\
&$\gamma=0.6$ & 48.55 & \ci{$-$0.58}{$-$1.12}{+0.03} &53.01\\
&$\gamma=0.8$ & 48.05 & \ci{$-$1.08}{$-$1.62}{$-$0.46} &52.93\\
\bottomrule
\end{tabular}}

\end{table}

\textbf{Change Tokens and Folding.}
As shown in \cref{tab:coverfold-ablation}(a), removing change tokens and folding reduces mIoU by 1.39 and 2.73, respectively.
This suggests that temporal changes carry critical but incomplete evidence, while folding preserves discarded content at the risk of blurring fine details.
The two mechanisms are therefore complementary in CoverFold.

With TEC, performance improves substantially across all configurations.
Notably, removing folding now slightly outperforms removing change tokens (52.57 vs. 52.37), reversing their ordering without TEC (46.40 vs. 47.74).
This shift suggests TEC's ability to recover discarded evidence directly through LLM attention, reducing reliance on folding, while explicit change protection remains valuable for retaining transient visual cues.
However, folding preserves discarded evidence by merging features, potentially blurring details, whereas TEC lets retained states attend directly to the original discarded features.
The two mechanisms therefore provide complementary paths for preserving evidence, with their combination achieving the highest mIoU (53.05).

\textbf{Coverage and Allocation.}
As shown in \cref{tab:coverfold-ablation}(a), uniform selection performs comparably to the coverage one (48.53 vs. 49.13), suggesting that the precise coverage strategy is less important.
However, retaining a few coverage-folding tokens is crucial under tight budgets to preserve evidence beyond changing regions.
Allocating 80\% of the budget to change tokens reduces mIoU by 1.08 (\cref{tab:coverfold-ablation}(b)), while TEC mitigates this sensitivity, maintaining 52.64--53.05 mIoU across allocations.

\subsection{Discussion}
\label{sec:discussion}
TEC consolidates discarded evidence into retained states, primarily textual timestamps, limiting direct transfer to other video representations.
However, visual-token-only consolidation also significantly improves mIoU (\cref{tab:tec-ablation}(b)), suggesting broader applicability.
Furthermore, TEC substantially increases LLM computation (\cref{tab:tec-ablation}(e)), but long-video inference is largely constrained by memory access.
By keeping the permanent KV cache compact, TEC incurs a much smaller latency increase than its computational overhead suggests (Appx.~E).
Additionally, KV quantization can further improve memory and inference efficiency without conflicting with TEC, and their combination achieves a favorable accuracy--efficiency balance as demonstrated in Appx.~E.
Finally, Appx.~D validates TEC in simulated real-world streaming, where questions can arrive before the video ends, demonstrating its potential for online temporal grounding.

\section{Conclusion}
We shift the focus of streaming video compression from selecting tokens to consolidating discarded visual evidence.
To this end, we propose \pruner, which protects temporal changes and folds discarded features into visual representatives, and \method, which  further integrates original discarded evidence into retained states, particularly timestamps, through early-layer LLM attention.
Together, they substantially improve temporal grounding under tight memory budgets, especially for queries involving small objects, without expanding the permanent KV cache or sacrificing streaming efficiency.
Our findings suggest a promising direction for building richer, more efficient video memory through evidence consolidation rather than token retention alone.

\label{sec:main-text-end}
\section*{LLM Usage Statement}
\addcontentsline{toc}{section}{LLM Usage Statement}
We developed the core research ideas and led the experimental design, execution, and collection of reported results.
OpenAI Codex assisted with literature organization, checking numerical consistency across tables and text, as well as language polishing, grammar and spelling correction.
We have reviewed all AI-assisted work and take responsibility for the final content of this work, including text, claims or artifacts produced with the aid of generative AI.

\par
{\small
\bibliographystyle{ieeenat_fullname}
\bibliography{main,mybib}

@article{duan2026novacov,
  title={Think in Sets for Streaming Video Token Compression},
  author={Duan, Moxu and Fu, Jingwen and Wang, Yuwang},
  journal={arXiv preprint arXiv:2608.01169},
  year={2026}
}

@inproceedings{zhang2026timelens,
  title={Timelens: Rethinking video temporal grounding with multimodal llms},
  author={Zhang, Jun and Wang, Teng and Ge, Yuying and Ge, Yixiao and Li, Xinhao and Wang, Limin},
  booktitle={Proceedings of the IEEE/CVF Conference on Computer Vision and Pattern Recognition},
  pages={10419--10429},
  year={2026}
}

@inproceedings{chen2024videollm,
  title={Videollm-online: Online video large language model for streaming video},
  author={Chen, Joya and Lv, Zhaoyang and Wu, Shiwei and Lin, Kevin Qinghong and Song, Chenan and Gao, Difei and Liu, Jia-Wei and Gao, Ziteng and Mao, Dongxing and Shou, Mike Zheng},
  booktitle={2024 IEEE/CVF Conference on Computer Vision and Pattern Recognition (CVPR)},
  pages={18407--18418},
  year={2024},
  organization={IEEE}
}

@inproceedings{huang2025online,
  title={Online video understanding: Ovbench and videochat-online},
  author={Huang, Zhenpeng and Li, Xinhao and Li, Jiaqi and Wang, Jing and Zeng, Xiangyu and Liang, Cheng and Wu, Tao and Chen, Xi and Li, Liang and Wang, Limin},
  booktitle={2025 IEEE/CVF Conference on Computer Vision and Pattern Recognition (CVPR)},
  pages={3328--3338},
  year={2025},
  organization={IEEE}
}

@inproceedings{sermanet2024robovqa,
  title={Robovqa: Multimodal long-horizon reasoning for robotics},
  author={Sermanet, Pierre and Ding, Tianli and Zhao, Jeffrey and Xia, Fei and Dwibedi, Debidatta and Gopalakrishnan, Keerthana and Chan, Christine and Dulac-Arnold, Gabriel and Maddineni, Sharath and Joshi, Nikhil J and others},
  booktitle={2024 IEEE International Conference on Robotics and Automation (ICRA)},
  pages={645--652},
  year={2024},
  organization={IEEE}
}

@inproceedings{yang2025egolife,
  title={Egolife: Towards egocentric life assistant},
  author={Yang, Jingkang and Liu, Shuai and Guo, Hongming and Dong, Yuhao and Zhang, Xiamengwei and Zhang, Sicheng and Wang, Pengyun and Zhou, Zitang and Xie, Binzhu and Wang, Ziyue and others},
  booktitle={2025 IEEE/CVF Conference on Computer Vision and Pattern Recognition (CVPR)},
  pages={28885--28900},
  year={2025},
  organization={IEEE}
}

@inproceedings{qian2025dispider,
  title={Dispider: Enabling video llms with active real-time interaction via disentangled perception, decision, and reaction},
  author={Qian, Rui and Ding, Shuangrui and Dong, Xiaoyi and Zhang, Pan and Zang, Yuhang and Cao, Yuhang and Lin, Dahua and Wang, Jiaqi},
  booktitle={2025 IEEE/CVF Conference on Computer Vision and Pattern Recognition (CVPR)},
  pages={24045--24055},
  year={2025},
  organization={IEEE}
}

@inproceedings{chen2026streamingtom,
  title={Streamingtom: Streaming token compression for efficient video understanding},
  author={Chen, Xueyi and Tao, Keda and Shao, Kele and Wang, Huan},
  booktitle={Proceedings of the IEEE/CVF Conference on Computer Vision and Pattern Recognition},
  pages={24675--24685},
  year={2026}
}

@inproceedings{wang2026accelerating,
  title={Accelerating streaming video large language models via hierarchical token compression},
  author={Wang, Yiyu and Liu, Xuyang and Gui, Xiyan and Lin, Xinying and Yang, Boxue and Liao, Chenfei and Chen, Tailai and Zhang, Linfeng},
  booktitle={Proceedings of the IEEE/CVF Conference on Computer Vision and Pattern Recognition},
  pages={18523--18533},
  year={2026}
}

@article{du2026unified,
  title={Unified spatiotemporal token compression for video-llms at ultra-low retention},
  author={Du, Junhao and Xue, Jialong and Li, Anqi and Dai, Jincheng and Lu, Guo},
  journal={arXiv preprint arXiv:2603.21957},
  year={2026}
}

@article{bai2025qwen3,
  title={Qwen3-vl technical report},
  author={Bai, Shuai and Cai, Yuxuan and Chen, Ruizhe and Chen, Keqin and Chen, Xionghui and Cheng, Zesen and Deng, Lianghao and Ding, Wei and Gao, Chang and Ge, Chunjiang and others},
  journal={arXiv preprint arXiv:2511.21631},
  year={2025}
}

@misc{qwen35blog,
    title = {Qwen3.5: Towards Native Multimodal Agents},
    url = {https://qwen.ai/blog?id=qwen3.5},
    author = {Qwen Team},
    month = {February},
    year = {2026}
}

@article{bolya2022token,
  title={Token merging: Your vit but faster},
  author={Bolya, Daniel and Fu, Cheng-Yang and Dai, Xiaoliang and Zhang, Peizhao and Feichtenhofer, Christoph and Hoffman, Judy},
  journal={arXiv preprint arXiv:2210.09461},
  year={2022}
}

@article{hooper2024kvquant,
  title={Kvquant: Towards 10 million context length llm inference with kv cache quantization},
  author={Hooper, Coleman and Kim, Sehoon and Mohammadzadeh, Hiva and Mahoney, Michael W and Shao, Yakun S and Keutzer, Kurt and Gholami, Amir},
  journal={Advances in Neural Information Processing Systems},
  volume={37},
  pages={1270--1303},
  year={2024}
}

@article{qin2025video,
  title={Video-xl-2: Towards very long-video understanding through task-aware kv sparsification},
  author={Qin, Minghao and Liu, Xiangrui and Liang, Zhengyang and Shu, Yan and Yuan, Huaying and Zhou, Juenjie and Xiao, Shitao and Zhao, Bo and Liu, Zheng},
  journal={arXiv preprint arXiv:2506.19225},
  year={2025}
}

@inproceedings{chen2024fastv,
  title={An image is worth 1/2 tokens after layer 2: Plug-and-play inference acceleration for large vision-language models},
  author={Chen, Liang and Zhao, Haozhe and Liu, Tianyu and Bai, Shuai and Lin, Junyang and Zhou, Chang and Chang, Baobao},
  booktitle={European Conference on Computer Vision},
  pages={19--35},
  year={2024},
  organization={Springer}
}

@article{xing2024pyramiddrop,
  title={Pyramiddrop: Accelerating your large vision-language models via pyramid visual redundancy reduction},
  author={Xing, Long and Huang, Qidong and Dong, Xiaoyi and Lu, Jiajie and Zhang, Pan and Zang, Yuhang and Cao, Yuhang and He, Conghui and Wang, Jiaqi and Wu, Feng and others},
  journal={arXiv preprint arXiv:2410.17247},
  year={2024}
}

@article{zhang2024sparsevlm,
  title={Sparsevlm: Visual token sparsification for efficient vision-language model inference},
  author={Zhang, Yuan and Fan, Chun-Kai and Ma, Junpeng and Zheng, Wenzhao and Huang, Tao and Cheng, Kuan and Gudovskiy, Denis and Okuno, Tomoyuki and Nakata, Yohei and Keutzer, Kurt and others},
  journal={arXiv preprint arXiv:2410.04417},
  year={2024}
}

@inproceedings{alvar2025divprune,
  title={Divprune: Diversity-based visual token pruning for large multimodal models},
  author={Alvar, Saeed Ranjbar and Singh, Gursimran and Akbari, Mohammad and Zhang, Yong},
  booktitle={2025 IEEE/CVF Conference on Computer Vision and Pattern Recognition (CVPR)},
  pages={9392--9401},
  year={2025},
  organization={IEEE}
}

@inproceedings{cho2026floc,
  title={Floc: Facility location-based efficient visual token compression for long video understanding},
  author={Cho, Janghoon and Lee, Jungsoo and Hayat, Munawar and Hwang, Kyuwoong and Porikli, Fatih and Choi, Sungha},
  booktitle={International Conference on Learning Representations},
  volume={2026},
  pages={102973--102996},
  year={2026}
}

@inproceedings{wen2025stop,
  title={Stop looking for “important tokens” in multimodal language models: Duplication matters more},
  author={Wen, Zichen and Gao, Yifeng and Wang, Shaobo and Zhang, Junyuan and Zhang, Qintong and Li, Weijia and He, Conghui and Zhang, Linfeng},
  booktitle={Proceedings of the 2025 Conference on Empirical Methods in Natural Language Processing},
  pages={9972--9991},
  year={2025}
}

@inproceedings{zhang2025beyond,
  title={Beyond text-visual attention: Exploiting visual cues for effective token pruning in vlms},
  author={Zhang, Qizhe and Cheng, Aosong and Lu, Ming and Zhang, Renrui and Zhuo, Zhiyong and Cao, Jiajun and Guo, Shaobo and She, Qi and Zhang, Shanghang},
  booktitle={2025 IEEE/CVF International Conference on Computer Vision (ICCV)},
  pages={20857--20867},
  year={2025},
  organization={IEEE}
}

@inproceedings{zhu2026speaks,
  title={Who Speaks for the Pruned? Visual Token Pruning as Coverage Optimization},
  author={Zhu, Qingchan and You, Weihang and Jiang, Hanqi and Yang, Changdi and Liu, Tianming and Yuan, Geng},
  booktitle={Proceedings of the 2026 Conference on Empirical Methods in Natural Language Processing},
  pages={8018--8037},
  year={2026}
}

@inproceedings{yao2025timechat,
  title={Timechat-online: 80\% visual tokens are naturally redundant in streaming videos},
  author={Yao, Linli and Li, Yicheng and Wei, Yuancheng and Li, Lei and Ren, Shuhuai and Liu, Yuanxin and Ouyang, Kun and Wang, Lean and Li, Shicheng and Li, Sida and others},
  booktitle={Proceedings of the 33rd ACM International Conference on Multimedia},
  pages={10807--10816},
  year={2025}
}

@inproceedings{shang2025llava,
  title={Llava-prumerge: Adaptive token reduction for efficient large multimodal models},
  author={Shang, Yuzhang and Cai, Mu and Xu, Bingxin and Lee, Yong Jae and Yan, Yan},
  booktitle={2025 IEEE/CVF International Conference on Computer Vision (ICCV)},
  pages={22857--22867},
  year={2025},
  organization={IEEE}
}

@inproceedings{yang2025visionzip,
  title={Visionzip: Longer is better but not necessary in vision language models},
  author={Yang, Senqiao and Chen, Yukang and Tian, Zhuotao and Wang, Chengyao and Li, Jingyao and Yu, Bei and Jia, Jiaya},
  booktitle={2025 IEEE/CVF Conference on Computer Vision and Pattern Recognition (CVPR)},
  pages={19792--19802},
  year={2025},
  organization={IEEE}
}

@inproceedings{tao2025dycoke,
  title={Dycoke: Dynamic compression of tokens for fast video large language models},
  author={Tao, Keda and Qin, Can and You, Haoxuan and Sui, Yang and Wang, Huan},
  booktitle={2025 IEEE/CVF Conference on Computer Vision and Pattern Recognition (CVPR)},
  pages={18992--19001},
  year={2025},
  organization={IEEE}
}

@inproceedings{fu2025framefusion,
  title={Framefusion: Combining similarity and importance for video token reduction on large vision language models},
  author={Fu, Tianyu and Liu, Tengxuan and Han, Qinghao and Dai, Guohao and Yan, Shengen and Yang, Huazhong and Ning, Xuefei and Wang, Yu},
  booktitle={2025 IEEE/CVF International Conference on Computer Vision (ICCV)},
  pages={22654--22663},
  year={2025},
  organization={IEEE}
}

@inproceedings{huang2025prunevid,
  title={Prunevid: Visual token pruning for efficient video large language models},
  author={Huang, Xiaohu and Zhou, Hao and Han, Kai},
  booktitle={Findings of the Association for Computational Linguistics: ACL 2025},
  pages={19959--19973},
  year={2025}
}

@article{shao2026holitom,
  title={Holitom: Holistic token merging for fast video large language models},
  author={Shao, Kele and Tao, Keda and Qin, Can and You, Haoxuan and Sui, Yang and Wang, Huan},
  journal={Advances in Neural Information Processing Systems},
  volume={38},
  pages={135547--135570},
  year={2026}
}

@article{kim2026infinipot,
  title={Infinipot-v: Memory-constrained kv cache compression for streaming video understanding},
  author={Kim, Minsoo and Shim, Kyuhong and Choi, Jungwook and Chang, Simyung},
  journal={Advances in Neural Information Processing Systems},
  volume={38},
  pages={138983--139013},
  year={2026}
}

@inproceedings{lin2025boosting,
  title={Boosting multimodal large language models with visual tokens withdrawal for rapid inference},
  author={Lin, Zhihang and Lin, Mingbao and Lin, Luxi and Ji, Rongrong},
  booktitle={Proceedings of the AAAI Conference on Artificial Intelligence},
  volume={39},
  number={5},
  pages={5334--5342},
  year={2025}
}

@inproceedings{deniz2025vision,
  title={Vision Token Reduction via Attention-Driven Self-Compression for Efficient Multimodal Large Language Models},
  author={Deniz, Omer Faruk and Mao, Ruiyu and Li, Ruochen and Tian, Yapeng and Khan, Latifur},
  booktitle={2025 IEEE International Conference on Big Data (BigData)},
  pages={4544--4551},
  year={2025},
  organization={IEEE}
}

@article{ou2026vision,
  title={When Vision Becomes Text: Visual Token Pruning via Cross-Modal Residual Guidance in VLMs},
  author={Ou, Congyang and Song, Ruike and Zhou, Yang and Sun, Libo and Zhang, Haokui and Luo, Zhenbo},
  journal={arXiv preprint arXiv:2608.10489},
  year={2026}
}

@article{chen2024sepllm,
  title={Sepllm: Accelerate large language models by compressing one segment into one separator},
  author={Chen, Guoxuan and Shi, Han and Li, Jiawei and Gao, Yihang and Ren, Xiaozhe and Chen, Yimeng and Jiang, Xin and Li, Zhenguo and Liu, Weiyang and Huang, Chao},
  journal={arXiv preprint arXiv:2412.12094},
  year={2024}
}

@inproceedings{wang2025seeing,
  title={Seeing more, saying more: Lightweight language experts are dynamic video token compressors},
  author={Wang, Xiangchen and Zhang, Jinrui and Wang, Teng and Zhang, Haigang and Zheng, Feng},
  booktitle={Proceedings of the 2025 Conference on Empirical Methods in Natural Language Processing},
  pages={541--558},
  year={2025}
}

@inproceedings{ye2025voco,
  title={Voco-llama: Towards vision compression with large language models},
  author={Ye, Xubing and Gan, Yukang and Huang, Xiaoke and Ge, Yixiao and Tang, Yansong},
  booktitle={2025 IEEE/CVF Conference on Computer Vision and Pattern Recognition (CVPR)},
  pages={29836--29846},
  year={2025},
  organization={IEEE}
}

@article{zheng2026visco,
  title={VisCo: Leveraging Large Language Models as Intrinsic Encoders for Visual Token Compression},
  author={Zheng, Yupeng and Zou, Kai and Liu, Bin and Yu, Nenghai},
  journal={arXiv preprint arXiv:2607.12756},
  year={2026}
}

@inproceedings{guo2025vtg,
  title={Vtg-llm: Integrating timestamp knowledge into video llms for enhanced video temporal grounding},
  author={Guo, Yongxin and Liu, Jingyu and Li, Mingda and Cheng, Dingxin and Tang, Xiaoying and Sui, Dianbo and Liu, Qingbin and Chen, Xi and Zhao, Kevin},
  booktitle={Proceedings of the AAAI conference on artificial intelligence},
  volume={39},
  number={3},
  pages={3302--3310},
  year={2025}
}

@inproceedings{huang2024lita,
  title={Lita: Language instructed temporal-localization assistant},
  author={Huang, De-An and Liao, Shijia and Radhakrishnan, Subhashree and Yin, Hongxu and Molchanov, Pavlo and Yu, Zhiding and Kautz, Jan},
  booktitle={European Conference on Computer Vision},
  pages={202--218},
  year={2024},
  organization={Springer}
}

@inproceedings{zeng2025timesuite,
  title={Timesuite: Improving mllms for long video understanding via grounded tuning},
  author={Zeng, Xiangyu and Li, Kunchang and Wang, Chenting and Li, Xinhao and Jiang, Tianxiang and Yan, Ziang and Li, Songze and Shi, Yansong and Yue, Zhengrong and Wang, Yi and others},
  booktitle={International Conference on Learning Representations},
  volume={2025},
  pages={38057--38081},
  year={2025}
}

@article{chen2024timemarker,
  title={Timemarker: A versatile video-llm for long and short video understanding with superior temporal localization ability},
  author={Chen, Shimin and Lan, Xiaohan and Yuan, Yitian and Jie, Zequn and Ma, Lin},
  journal={arXiv preprint arXiv:2411.18211},
  year={2024}
}

@inproceedings{li2026keeping,
  title={Keeping the evidence chain: Semantic evidence allocation for training-free token pruning in video temporal grounding},
  author={Li, Jiaqi and Zheng, Shuntian and Shen, Yixian and Huang, Jia-Hong and Lu, Xiaoman and Ni, Minzhe and Guan, Yu},
  booktitle={European Conference on Computer Vision},
  pages={549--567},
  year={2026},
  organization={Springer}
}

@article{fan2026flashvid,
  title={Flashvid: Efficient video large language models via training-free tree-based spatiotemporal token merging},
  author={Fan, Ziyang and Chen, Keyu and Xing, Ruilong and Li, Yulin and Jiang, Li and Tian, Zhuotao},
  journal={arXiv preprint arXiv:2602.08024},
  year={2026}
}

@article{shen2026fastvid,
  title={Fastvid: Dynamic density pruning for fast video large language models},
  author={Shen, Leqi and Gong, Guoqiang and He, Tao and Zhang, Yifeng and Liu, Pengzhang and Zhao, Sicheng and others},
  journal={Advances in Neural Information Processing Systems},
  volume={38},
  pages={123553--123581},
  year={2026}
}

@inproceedings{ren2024timechat,
  title={Timechat: A time-sensitive multimodal large language model for long video understanding},
  author={Ren, Shuhuai and Yao, Linli and Li, Shicheng and Sun, Xu and Hou, Lu},
  booktitle={2024 IEEE/CVF Conference on Computer Vision and Pattern Recognition (CVPR)},
  pages={14313--14323},
  year={2024},
  organization={IEEE}
}

@inproceedings{krishna2017dense,
  title={Dense-captioning events in videos},
  author={Krishna, Ranjay and Hata, Kenji and Ren, Frederic and Fei-Fei, Li and Carlos Niebles, Juan},
  booktitle={Proceedings of the IEEE international conference on computer vision},
  pages={706--715},
  year={2017}
}

@inproceedings{gao2017charades,
  title={Tall: Temporal activity localization via language query},
  author={Gao, Jiyang and Sun, Chen and Yang, Zhenheng and Nevatia, Ram},
  booktitle={2017 IEEE International Conference on Computer Vision (ICCV)},
  pages={5277--5285},
  year={2017},
  organization={IEEE}
}

@inproceedings{fu2025video,
  title={Video-mme: The first-ever comprehensive evaluation benchmark of multi-modal llms in video analysis},
  author={Fu, Chaoyou and Dai, Yuhan and Luo, Yongdong and Li, Lei and Ren, Shuhuai and Zhang, Renrui and Wang, Zihan and Zhou, Chenyu and Shen, Yunhang and Zhang, Mengdan and others},
  booktitle={2025 IEEE/CVF Conference on Computer Vision and Pattern Recognition (CVPR)},
  pages={24108--24118},
  year={2025},
  organization={IEEE}
}

@article{wu2024longvideobench,
  title={Longvideobench: A benchmark for long-context interleaved video-language understanding},
  author={Wu, Haoning and Li, Dongxu and Chen, Bei and Li, Junnan},
  journal={Advances in Neural Information Processing Systems},
  volume={37},
  pages={28828--28857},
  year={2024}
}

@inproceedings{di2025streaming,
  title={Streaming video question-answering with in-context video kv-cache retrieval},
  author={Di, Shangzhe and Yu, Zhelun and Zhang, Guanghao and Li, Haoyuan and Cheng, Hao and Li, Bolin and He, Wanggui and Shu, Fangxun and Jiang, Hao},
  booktitle={International Conference on Learning Representations},
  volume={2025},
  pages={42115--42127},
  year={2025}
}

@inproceedings{ji2020action,
  title={Action genome: Actions as compositions of spatio-temporal scene graphs},
  author={Ji, Jingwei and Krishna, Ranjay and Fei-Fei, Li and Niebles, Juan Carlos},
  booktitle={2020 IEEE/CVF Conference on Computer Vision and Pattern Recognition (CVPR)},
  pages={10233--10244},
  year={2020},
  organization={IEEE}
}

@article{wang2026era,
  title={ERA: Entropy-Guided Visual Token Pruning with Rectified Attention for Efficient MLLMs},
  author={Wang, Yuhao and Qiao, Mu and Diao, Haiwen and Zhuge, Yunzhi and Zhang, Pingping and Zhang, Xindong and Zhang, Lei and Lu, Huchuan},
  journal={arXiv preprint arXiv:2606.31982},
  year={2026}
}

@article{lv2026vprune,
  title={VPRune: Efficient Training-free Pre-LLM Visual Token Pruning},
  author={Lv, Guangchuan and Shi, Dianxing and Fu, Dingjie},
  journal={arXiv preprint arXiv:2609.24485},
  year={2026}
}

@article{mangalam2023egoschema,
  title={Egoschema: A diagnostic benchmark for very long-form video language understanding},
  author={Mangalam, Karttikeya and Akshulakov, Raiymbek and Malik, Jitendra},
  journal={Advances in Neural Information Processing Systems},
  volume={36},
  pages={46212--46244},
  year={2023}
}

@article{seo2026natural,
  title={Natural-Language Temporal Grounding in Hour-Long Videos is a Search Problem: A Benchmark and Empirical Decomposition},
  author={Seo, Sukmin and Kim, Geewook},
  journal={arXiv preprint arXiv:2606.12300},
  year={2026}
}

@article{zhang2026covist,
  title={CoViST: Visual Token Compression via Composable States},
  author={Zhang, Qi and Meng, Xiandong and Wang, Ronggang and Ma, Siwei},
  journal={arXiv preprint arXiv:2609.33397},
  year={2026}
}

@inproceedings{niu2025ovo,
  title={Ovo-bench: How far is your video-llms from real-world online video understanding?},
  author={Niu, Junbo and Li, Yifei and Miao, Ziyang and Ge, Chunjiang and Zhou, Yuanhang and He, Qihao and Dong, Xiaoyi and Duan, Haodong and Ding, Shuangrui and Qian, Rui and others},
  booktitle={2025 IEEE/CVF Conference on Computer Vision and Pattern Recognition (CVPR)},
  pages={18902--18913},
  year={2025},
  organization={IEEE}
}

@inproceedings{zhang2026hermes,
  title={Hermes: Kv cache as hierarchical memory for efficient streaming video understanding},
  author={Zhang, Haowei and Yang, Shudong and Fu, Jinlan and Ng, See Kiong and Qiu, Xipeng},
  booktitle={Proceedings of the 64th Annual Meeting of the Association for Computational Linguistics (Volume 1: Long Papers)},
  pages={8411--8430},
  year={2026}
}

@article{xie2026fluxmem,
  title={Fluxmem: Adaptive hierarchical memory for streaming video understanding},
  author={Xie, Yiweng and He, Bo and Wang, Junke and Zheng, Xiangyu and Ye, Ziyi and Wu, Zuxuan},
  journal={arXiv preprint arXiv:2603.02096},
  year={2026}
}

@article{dong2026objectstream,
  title={ObjectStream: Latent Objects as Memory Anchors for Streaming Video Understanding},
  author={Dong, Mingkang and Pu, Muxin and Li, Jie and Guo, Bohan and Chen, Songruo and Ren, Bin and Zheng, Xu and Zhao, Chen and Qian, Tianwen and Elhoseiny, Mohamed and others},
  journal={arXiv preprint arXiv:2607.28312},
  year={2026}
}

@article{hao2026shallowstream,
  title={ShallowStream: Index Shallow then Answer Deep for Streaming Video Understanding},
  author={Hao, Jitai and Yang, Ke and Yan, Di and Liu, Fan and Huang, Qiang and Yu, Jun},
  journal={arXiv preprint arXiv:2609.02780},
  year={2026}
}

@article{song2026towards,
  title={Towards a Dynamic and Fixed-budget Memory Bank for Efficient Streaming Video Understanding},
  author={Song, Baiyang and Lin, Yuli and Wu, Qiong and Chen, Tao and Peng, Jun and Chen, Xiao and Zhou, Yiyi and Ji, Rongrong},
  journal={arXiv preprint arXiv:2606.25658},
  year={2026}
}

@article{xiao2026mukv,
  title={Mukv: Multi-grained kv cache compression for long streaming video question-answering},
  author={Xiao, Junbin and Chen, Jiajun and Sun, Tianxiang and Yang, Xun and Yao, Angela},
  journal={arXiv preprint arXiv:2605.22269},
  year={2026}
}
}

\clearpage
\phantomsection
\label{sec:supplement-start}
\appendix

\section{Implementation Details}

\subsection{\pruner Implementation}
\label{appx:coverfold_impl}
The norm restoration factor in Eq.~(3) of the main paper is computed as
\begin{equation}
\rho_k =
\frac{\|x_k\|+\sum_{j\in\mathcal A_k}w_{jk}\|x_j\|}
{1+\sum_{j\in\mathcal A_k}w_{jk}}.
\end{equation}
For Qwen3-VL, the same folding assignments and weights are applied to DeepStack features. Folded features occupy retained visual positions, while discarded tokens preserve their original features, DeepStack features, and positions for \method. The pruning-only baseline uses identical selected and folded tokens to isolate the contribution of \method.

\subsection{\method Implementation}

We implement evidence transfer using a causal attention mask and a temporary KV cache, following Equations~(5) and~(6) of the main paper.

\textbf{Attention Mask.}
TEC allows retained tokens to read recently dropped evidence while preserving the original causal token order. Let $\Pset=\Sset\cup\Rset$ denote all retained positions, and $\Dset_A$ the active dropped positions, comprising the current input and those still available from the recent $W$ inputs. The attention mask at layer $\ell$ is
\begin{equation}
\begin{aligned}
M_{ij}^{\ell}
&=\mathbf{1}[j\preceq i]\,
\mathbf{1}[j\in\mathcal K_i^\ell],\\
\mathcal K_i^\ell
&=\Pset\cup
\begin{cases}
\Dset_A, & i\in\Pset,\ 1\le\ell\le b,\\
\Dset_A, & i\in\Dset_t,\ \ell\le b,\\
\varnothing, & \text{otherwise}.
\end{cases}
\end{aligned}
\label{eq:mask}
\end{equation}
where $\preceq$ enforces causal order and $\mathcal K_i^\ell$ denotes the keys accessible to query $i$. Retained tokens can read active dropped evidence in the first $b$ layers, while current dropped tokens attend to retained history and active dropped tokens through layer $b$. All other attention follows standard pruning. The window $W$ is enforced through temporary KV eviction, and dropped tokens do not proceed beyond layer $b$.

\subsection{Overall Procedure}
Algorithm~\ref{alg} summarizes streaming inference. CoverFold selects and optionally folds visual features before LLM processing. TEC then temporarily includes dropped tokens in early-layer attention, while only retained tokens complete all $L$ layers and enter long-term memory. Dropped KV states retain their original positions until their window expires.

\begin{algorithm}[t]
\caption{Streaming Compression with \pruner and \method}
\label{alg}
\begin{algorithmic}[1]
\Require Frozen LLM with $L$ layers; pruner; $b,W$
\State Initialize long-term KV memory $\mathcal M$ and temporary KV cache $\mathcal C$
\For{each incoming visual input $t$}
    \State Evict from $\mathcal C$ dropped KV older than $t-W+1$
    \State Obtain $\Sset_t,X_t$; select $\Rset_t,\Dset_t$ by CoverFold
    \State Fold $\Dset_t$ into $\Rset_t$
    \State Process $\Sset_t,\Rset_t,\Dset_t$ through layers $1{:}b$ under \cref{eq:mask}
    \State Process $\Sset_t,\Rset_t$ through layers $b{+}1{:}L$
    \State Append retained KV states to $\mathcal M$
    \State Append dropped KV states to $\mathcal C$
\EndFor
\State Clear $\mathcal C$ before question answering
\State \Return $\mathcal M$
\end{algorithmic}
\end{algorithm}

\section{Evaluation Details}

\subsection{Benchmarks}

\textbf{Charades-STA.}
Charades-STA~\cite{gao2017charades} evaluates language-guided localization of everyday activities in indoor videos. We use its complete test split of 3,720 query--moment pairs from 1,334 videos. Videos average approximately 30 seconds, with annotated events lasting around 8 seconds. The frequent overlap of activities and visually similar indoor scenes requires precise temporal boundaries and preservation of subtle action changes.

\textbf{ActivityNet Captions.}
ActivityNet Captions~\cite{krishna2017dense} contains untrimmed videos spanning diverse activities and complex events. Unlike Charades-STA, its videos average roughly two minutes, with longer annotated moments and greater variation in scenes and event durations. We evaluate a fixed subset of 600 videos from the \texttt{val\_2} split, comprising 2,142 query--moment pairs. This benchmark tests whether compressed visual histories retain sufficient context to localize events over longer temporal spans.

\textbf{Video-MME.}
Video-MME~\cite{fu2025video} assesses video understanding across six major domains and 30 subcategories, covering short, medium, and long videos ranging from 11 seconds to one hour. It contains 900 videos totaling approximately 254 hours, with 2,700 human-annotated multiple-choice questions. The questions require recognition, information retrieval, and temporal reasoning across diverse content. We use the subtitle-free setting to evaluate how well compressed visual memory supports question answering without textual descriptions of the video.

\textbf{LongVideoBench.}
LongVideoBench~\cite{wu2024longvideobench} emphasizes long-context understanding, particularly referring reasoning that requires identifying and connecting details across different video moments. The full benchmark covers 17 question categories and videos up to one hour long. We evaluate its validation split of 1,337 multiple-choice questions over 753 videos, without subtitles. Its emphasis on fine-grained details and cross-moment dependencies tests whether compression preserves evidence that may become relevant much later.

\textbf{OVO-Bench.}
OVO-Bench~\cite{niu2025ovo} evaluates 12 tasks across real-time visual perception, backward tracing, and forward active responding.
The released annotations used here contain 1,640 questions over 644 videos.
Forward-response questions are probed at multiple time points, yielding 1,567 such probes and 3,035 scored queries in total.
Tasks include recognizing current content, recalling past events, and tracking counts or procedural steps as new frames arrive.
This tests whether compressed memory remains useful throughout the stream; the protocol and results are in \cref{sec:ovobench}.

\textbf{Small-Object Grounding.}
Visual token pruning may discard small objects, while merging may dilute their features with surrounding content. 
To examine whether compression disproportionately affects such fine-grained evidence, we construct a small-object subset of Charades-STA using bounding-box annotations from Action Genome~\cite{ji2020action}.
We match queries to 35 object categories through case-insensitive whole-word matching with a small synonym lexicon.
For each mentioned category, we compute the median visible bounding-box area across the video's annotated frames, normalized by frame area. 
The median provides a robust size estimate, while taking the minimum across matched categories focuses on the smallest object referenced by each query.
Categories without visible annotations are excluded.
Of 3,720 test queries, 2,989 have measurable object sizes; the remaining 731 either mention no matched category (576) or lack corresponding visible boxes (155).
We select 854 queries (23.0\% of the test set) with object areas below 1\%, roughly the spatial scale of a single visual token.
This threshold was fixed from the size distribution before examining model performance.

\subsection{Evaluation Metrics}
\label{sec:eval-metrics}

For temporal grounding, we report mean temporal intersection-over-union (mIoU), averaging the overlap between predicted and ground-truth intervals across queries~\cite{ren2024timechat,li2026keeping}.
For small-object grounding, we additionally report subset mIoU and its absolute drop from the uncompressed model on the same queries, $\Delta=\mathrm{mIoU}_{\mathrm{all}}-\mathrm{mIoU}_{\mathrm{method}}$; lower is better. Other accuracy-difference columns subtract the stated reference from the row.
For question answering, we report multiple-choice accuracy following the respective benchmark protocols~\cite{fu2025video,wu2024longvideobench}.

To assess the statistical reliability, we compute 95\% confidence intervals using 5,000 video-clustered bootstrap resamples. 
Each resample draws videos with replacement and includes all their associated queries, accounting for correlations among queries from the same video. 
For method comparisons, we resample the same videos for both methods and compute confidence intervals on their paired score differences, allowing us to assess whether observed improvements are statistically significant.

For efficiency in the main table, we directly time each method's complete per-step compression and LLM execution.
The measured interval includes token selection, folding and TEC where applicable, LLM inference, and KV-cache updates and eviction; vision encoding is excluded.
Within each backbone, methods receive identical sampled inputs and use the same model precision.
We report mean step latency (ms) over the final minute of 10-minute and one-hour streams, and memory usage (GiB) at the final step.
For uncompressed Qwen3-VL runs requiring KV offloading, reported memory includes host storage rather than GPU memory alone.

TFLOPs denote total floating-point operations over the one-hour stream, rather than throughput.
We sum matrix-operation counts across steps and layers, counting each multiply-add as two operations and including both query--key and attention--value products.
The counts use each step's input and cache sizes, with TEC's discarded tokens counted only through their processing depth $b$.
Analytical counts for uncompressed inference are marked with $^\dagger$.

\subsection{Protocols}

\textbf{Qwen3-VL and Qwen3.5.}
We evaluate frozen Qwen3-VL-4B/8B~\cite{bai2025qwen3}, whose 36-layer full-attention decoders maintain growing KV histories, and Qwen3.5-9B~\cite{qwen35blog}, which uses 24 Gated DeltaNet linear-attention layers and eight full-attention layers.
By summarizing most historical information in fixed-size recurrent states, Qwen3.5 reduces long-context attention costs, achieving roughly $2\times$ lower per-step latency on one-hour streams despite its larger size.
We use the Instruct checkpoint for efficiency consideration.

\textbf{Temporal Grounding Protocol.}
Videos are sampled at 2\,fps (up to 768 frames), with Qwen3-VL grouping two consecutive frames into a timestamped step containing at most 196 visual tokens.
Each step is compressed causally, and the query is provided only after video encoding, ensuring that compression cannot exploit future questions.

For Qwen3-VL, we use the SemVID prompt~\cite{li2026keeping}:
\begin{quote}\small
Give you a textual query: \{query\}.\\
When does the described content occur in the video?\\
Please return the final timestamps in seconds directly in one sentence.
\end{quote}
For Qwen3.5, we use the TimeChat prompt~\cite{ren2024timechat}, as the SemVID prompt often yields incomplete responses:
\begin{quote}\small
Localize the visual content described by the given textual query `\{query\}' in the video, and output the start and end timestamps in seconds.
\end{quote}
Prompt choices are based on uncompressed performance on training videos.
We allow up to 64 generated tokens and assign zero IoU to unparsable predictions.

\textbf{Question Answering Protocol.}
Following ReKV~\cite{di2025streaming}, videos are sampled at 0.5\,fps and encoded sequentially within a 15,000-token attention window.
Historical KV states are stored as timestamped steps, each represented for retrieval by its mean key before rotary encoding.
When a question arrives, each LLM layer retrieves the 64 most relevant steps in temporal order.
The question and answer options are accompanied by the instruction "Only give the best option." and the generation prefix "Best option: ".
TEC's temporary dropped-token KV is excluded from stored steps and retrieval, preserving the same long-term token budget as the corresponding pruning baseline.

\textbf{Compression Protocol.}
For fair comparison, all methods use identical queries, annotations, visual features, and DeepStack handling.
We evaluate 12.5\% and 25\% retention, counting native timestamps and temporal markers within the total token budget.
At 12.5\% retention, timestamps occupy 49.7\%, 65.8\%, and 40.8\% of retained positions on Charades-STA, ActivityNet, and Video-MME, respectively.
This substantial overhead motivates consolidating discarded visual evidence into existing timestamp states rather than allocating additional long-term KV positions.

\subsection{Baselines}

\textbf{Streaming Visual Compression.}
We compare DTD~\cite{yao2025timechat} and StreamingTOM~\cite{chen2026streamingtom}, which exploit temporal changes and static--dynamic redundancy, alongside STC-Pruner~\cite{wang2026accelerating} and NovaCov~\cite{duan2026novacov}, which use historical information to identify novel or representative features.
DTD's released budget-allocation rule assumes a known video length but does not access future frames.

\textbf{Per-Step Visual Compression.}
We adapt VisionZip~\cite{yang2025visionzip}, ToMe~\cite{bolya2022token}, DivPrune~\cite{alvar2025divprune}, and FastVID~\cite{shen2026fastvid} to compress each incoming step independently.
These methods span attention-based selection, diversity maximization, and similarity-based merging.
In particular, FastVID's video-level segmentation is restricted to the current step to prevent access to future content.
These adaptations compare compression mechanisms under streaming constraints.

\textbf{KV Compression and Non-Streaming References.}
Unlike pre-LLM visual pruning, InfiniPot-V~\cite{kim2026infinipot} processes incoming tokens through the full LLM before compressing their KV states using temporal redundancy and value importance.
HERMES~\cite{zhang2026hermes} similarly processes all tokens through the full decoder, retaining recent entries in shallow layers, pseudo-question-relevant entries in deeper layers, and averaging evicted states into summaries.
In contrast, \method integrates original discarded features into temporally aligned timestamps through early-layer attention, avoiding full-depth processing.
We adapt both methods to Qwen3-VL's rotary positions and ReKV's retrieval protocol while matching the retained-position budget.
We do not combine them with \method because their history-wide KV eviction occurs after full-depth processing, potentially transferring evidence from earlier steps into unrelated timestamps.
FlashVID~\cite{fan2026flashvid} and SemVID~\cite{li2026keeping} provide non-streaming references: the former exploits complete-video processing and query-dependent late-layer selection, while the latter uses the question and future visual context.
Their results are therefore not directly subject to the same causal, query-agnostic constraints.

\textbf{Early-Layer Withdrawal.}
VTW~\cite{lin2025boosting} withdraws all visual tokens after shallow-layer information transfer to text, assuming instructions are available during visual processing.
Unlike VTW, \method complements with pruner, preserves selected visual tokens and lets subsequent timestamps absorb discarded evidence before eviction, maintaining its temporal association.

\textbf{Other Streaming Memory Methods.}
SepLLM~\cite{chen2024sepllm} preserves separator tokens as memory carriers, whereas Video-XL-2~\cite{qin2025video} uses timestamp KV alongside stored visual states and query-dependent retrieval.
FluxMem~\cite{xie2026fluxmem}, ObjectStream~\cite{dong2026objectstream}, and CausalMem~\cite{song2026towards} employ adaptive or global memory allocation rather than fixed per-step retention.
ShallowStream~\cite{hao2026shallowstream} and MuKV~\cite{xiao2026mukv} rely on query-time processing or retrieval, introducing different storage and latency trade-offs.
Their published results are not directly comparable under our fixed KV budgets without adapting their protocols.
Unlike these systems, \method focuses on preserving discarded evidence within existing KV positions without requiring future questions or re-encoding.

\section{Additional Results}
\label{sec:additional-results}
\begin{table}[t]
\caption{\textbf{Grounding at different IoU thresholds.} R@$u$: recall at temporal IoU $\geq u$.}
\label{tab:recall}
\centering\small
\setlength{\tabcolsep}{4pt}
\begin{tabular}{lcccc}
\toprule
Charades-STA & R@0.3 & R@0.5 & R@0.7 & mIoU\\
\midrule
All tokens & 79.65 & 65.86 & 40.70 & 55.99\\
Pruning, 12.5\% & 73.15 & 56.24 & 30.40 & 49.13\\
+ \method, 12.5\% & \textbf{76.10} & \textbf{62.12} & \textbf{36.94} & \textbf{53.05}\\
Pruning, 25\% & \textbf{78.47} & 62.45 & 35.65 & 53.57\\
+ \method, 25\% & 77.98 & \textbf{63.12} & \textbf{38.74} & \textbf{54.45}\\
\bottomrule
\end{tabular}

\end{table}

\textbf{Temporal Overlap.}
At 12.5\%, \method raises R@0.7 by 6.54 points, compared with 2.95 at R@0.3 (\cref{tab:recall}).
At 25\%, strict-overlap recall still improves although coarse recall falls slightly.
The improvement is therefore concentrated at stricter overlap thresholds, consistent with better cues for placing event boundaries rather than only detecting that an event occurs.
This complements the main paper's small-object analysis: preserving local visual evidence can improve both what is recognized and where its temporal extent is placed.

\begin{table}[t]
\caption{\textbf{Additional \pruner sensitivity.} $\Delta$: paired change in mIoU points from the default row in each group.}
\label{tab:sensitivity}
\centering
\centering\footnotesize
\setlength{\tabcolsep}{2.5pt}
\begin{tabular}{lcccc}
\toprule
Setting & 12.5\% & $\Delta$ [95\% CI] & 25\% & $\Delta$ [95\% CI]\\
\midrule
\multicolumn{5}{l}{\emph{Norm exponent $\kappa$}}\\
$\boldsymbol{\kappa=0.5}$ & \textbf{49.13} & ref. & \textbf{53.57} & ref.\\
$\kappa=1$ & 49.03 & \ci{$-$0.10}{$-$0.59}{+0.41} & 53.25 & \ci{$-$0.32}{$-$0.70}{+0.06}\\
$\kappa=2$ & 49.11 & \ci{$-$0.02}{$-$0.63}{+0.59} & 52.95 & \ci{$-$0.62}{$-$1.03}{$-$0.22}\\
\midrule
\multicolumn{5}{l}{\emph{Folding threshold $\tau$}}\\
$\boldsymbol{\tau=0}$ & \textbf{49.13} & ref. & 53.57 & ref.\\
$\tau=0.2$ & 48.94 & \ci{$-$0.19}{$-$0.55}{+0.15} & \textbf{53.64} & \ci{+0.07}{$-$0.23}{+0.36}\\
$\tau=0.4$ & 48.86 & \ci{$-$0.27}{$-$0.65}{+0.12} & 53.57 & \ci{+0.00}{$-$0.29}{+0.29}\\
$\tau=0.6$ & 48.23 & \ci{$-$0.90}{$-$1.43}{$-$0.34} & 53.54 & \ci{$-$0.03}{$-$0.33}{+0.28}\\
\bottomrule
\end{tabular}

\end{table}

\textbf{Additional Sensitivity Analysis.}
Moderate changes to the norm weights and folding threshold have paired intervals spanning zero (\cref{tab:sensitivity}).
A large norm exponent hurts at 25\%, while a strict folding threshold hurts at 12.5\%.
The former concentrates selection on high-norm features; the latter rejects more discarded evidence.
These results favor moderate weighting and permissive folding, rather than aggressively filtering the evidence before the LLM sees it.

\begin{table}[t]
\caption{\textbf{Additional retention budgets.} Shading adds \method.}
\label{tab:budgets-extra}
\centering\footnotesize
\setlength{\tabcolsep}{2.5pt}
\newcommand{\tecrow}{\rowcolor{gray!13}\quad + \method}
\resizebox{0.75\linewidth}{!}{%
\begin{tabular}{lcccc}
\toprule
Method & \shortstack{Charades-\\STA} & ActivityNet & Video-MME & LongVB\\
\midrule
All tokens & 55.99 & 42.81 & 66.81 & 61.11\\
\midrule
\multicolumn{5}{l}{\emph{One visual token per step}}\\
NovaCov & 27.90 & 35.62 & 45.52 & 46.82\\
\tecrow & 50.56 & 36.68 & 49.72 & 47.87\\
VisionZip & 21.05 & 32.07 & 43.33 & 47.05\\
\tecrow & 50.21 & 37.06 & \textbf{51.15} & 47.94\\
\pruner & 24.23 & 34.85 & 44.00 & 46.67\\
\tecrow & \textbf{50.58} & \textbf{37.63} & 50.86 & \textbf{48.24}\\
\midrule
\multicolumn{5}{l}{\emph{33\% retention}}\\
NovaCov & 53.01 & 42.47 & 65.56 & 59.31\\
\tecrow & 54.53 & 42.64 & \textbf{65.81} & 59.24\\
VisionZip & 53.73 & 42.37 & 64.19 & 58.12\\
\tecrow & 54.74 & 42.96 & 64.26 & 57.97\\
\pruner & 54.45 & 42.90 & 65.52 & \textbf{60.28}\\
\tecrow & \textbf{55.20} & \textbf{43.44} & 65.77 & \textbf{60.28}\\
\bottomrule
\end{tabular}}

\end{table}

\textbf{Additional Retention Budgets.}
As shown in \cref{tab:budgets-extra}, with just one visual token per step, TEC brings different pruners to similar accuracy despite their widely varying pruning-only performance.
This further shows that consolidating discarded evidence reduces dependence on token selection, as timestamps provide a shared path for preserving otherwise lost information.

\begin{table}[t]
\caption{\textbf{Visual evidence in stored text states.} In (b), visual KV stays fixed; text KV includes timestamps, markers, and prompt text.}
\label{tab:timestamp-memory}
\centering
\textbf{(a) Compare stored KV with uncompressed video}\par\smallskip
\centering\small
\setlength{\tabcolsep}{4pt}
\resizebox{0.72\linewidth}{!}{%
\begin{tabular}{lcccc}
\toprule
Cosine similarity & \multicolumn{2}{c}{Time text} & \multicolumn{2}{c}{Markers}\\
\cmidrule(lr){2-3}\cmidrule(lr){4-5}
Memory & Keys & Values & Keys & Values\\
\midrule
Pruning & 0.927 & 0.837 & 0.955 & 0.904\\
\method & \textbf{0.977} & \textbf{0.938} & \textbf{0.990} & \textbf{0.977}\\
Gap closed by \method & 69\% & 62\% & 77\% & 76\%\\
\midrule
Other video (full) & 0.902 & 0.750 & -- & --\\
\bottomrule
\end{tabular}}
\par\medskip
\textbf{(b) Swap text KV before answering}\par\smallskip
\centering\small
\setlength{\tabcolsep}{3pt}
\resizebox{0.65\linewidth}{!}{%
\begin{tabular}{lcc}
\toprule
Text KV source & mIoU & Change [95\% CI]\\
\midrule
\multicolumn{3}{l}{\emph{Keep \method's visual KV fixed}}\\
\method (unchanged) & \textbf{53.05} & ref.\\
Pruning & 49.30 & \ci{$-$3.75}{$-$4.53}{$-$2.94}\\
\midrule
\multicolumn{3}{l}{\emph{Keep pruning's visual KV fixed}}\\
Pruning (unchanged) & 49.13 & ref.\\
\method & 52.21 & \ci{+3.08}{+2.24}{+3.91}\\
Uncompressed & 52.48 & \ci{+3.35}{+2.48}{+4.21}\\
\bottomrule
\end{tabular}}

\end{table}

\textbf{Evidence Retention in Timestamps.}
As shown in \cref{tab:timestamp-memory}(a), we compare timestamp KV states from pruning and TEC against those from uncompressed processing of the same video, matching token positions and LLM layers.
TEC increases timestamp-value similarity from 0.837 to 0.938, indicating that its retained timestamp states more closely preserve the representations formed with full visual evidence.
In contrast, uncompressed timestamps at matching positions across different videos have a similarity of only 0.750, indicating that these states encode video-dependent information beyond their time labels.
Together, these results suggest that TEC better preserves visual information within the retained timestamp states.

To test whether these states affect the answer, \cref{tab:timestamp-memory}(b) swaps stored text KV between memories of the same video, keeping visual KV fixed.
Text includes timestamps, markers, and the prompt template.
Replacing TEC's text KV with pruning's lowers mIoU from 53.05 to 49.30; the reverse swap raises pruning from 49.13 to 52.21.
Thus, much of the gain can be removed or restored through text states alone.
Together with (a), this shows that useful visual evidence persists in stored text memory after its source tokens have been evicted.

\section{Online Temporal Grounding}
\label{sec:online-grounding}
\begin{table}[t]
\caption{\textbf{Grounding at different query delays.} mIoU; ExtremeWhenBench columns give minutes from event end to query. }
\label{tab:online-grounding}
\centering\footnotesize
\setlength{\tabcolsep}{3pt}
\newcommand{\onlineTEC}{\rowcolor{gray!10}\quad + \method}
\resizebox{\linewidth}{!}{%
\begin{tabular}{lcc|cc|ccccccc}
\toprule
\multirow{2}{*}{Method} & \multicolumn{2}{c}{Charades-STA} & \multicolumn{2}{c}{ActivityNet} & \multicolumn{7}{c}{ExtremeWhenBench: event-to-query gap (min)}\\
\cmidrule(lr){2-3}\cmidrule(lr){4-5}\cmidrule(lr){6-12}
& Random & End & Random & End & 0--1 & 1--5 & 5--15 & 15--30 & $>$30 & End & All\\
\midrule
All tokens & 55.95 & 55.99 & 45.70 & 42.81 & 21.43 & 16.48 & 12.93 & 10.39 & 4.93 & 8.26 & 13.29\\
\midrule
ToMe & 39.18 & 42.46& 40.57 & 40.48& 12.78 & 10.93 & 7.99 & 7.36 & 6.52 & 5.78 & 8.85\\
\onlineTEC & 51.80 & 52.32& 41.85& 41.34& 16.09 & 13.72 & 8.57 & 6.71 & 5.12 & 6.07 & 9.99\\
VisionZip & 38.27& 37.52& 36.37& 36.14& 10.15 & 8.87 & 6.03 & 5.40 & 4.87 & 4.08 & 6.82\\
\onlineTEC & 52.15 & 52.16& 40.68& 39.82& 15.74 & 12.03 & 8.83 & 6.58 & 4.86 & 5.57 & 9.50\\
DTD & 39.28& 39.07& 35.79& 34.82& 13.06 & 10.88 & 8.44 & \textbf{8.47} & 7.74 & 6.02 & 9.29\\
\onlineTEC & 53.26& 52.31& 40.49& 40.34& 16.83 & 13.76 & \textbf{10.16} & 8.46 & 7.93 & 7.56 & 11.22\\
NovaCov & 44.72& 44.02& 39.60& 39.13& 14.72 & 12.81 & 9.38 & 7.65 & 7.90 & 6.60 & 10.17\\
\onlineTEC & 52.11 & 52.98& 41.99 & 41.42& 17.59 & 15.52 & 9.28 & 7.75 & 7.39 & 7.30 & 11.36\\
\pruner & 44.15 & 49.13 & \textbf{42.20} & 40.76 & 13.17 & 11.70 & 8.81 & 7.47 & 7.06 & 6.40 & 9.41\\
\onlineTEC & \textbf{53.68} & \textbf{53.05} & 42.15 & \textbf{42.01} & \textbf{17.98} & \textbf{15.57} & 9.77 & 7.95 & \textbf{7.94} & \textbf{7.77} & \textbf{11.71}\\
\bottomrule
\end{tabular}}

\end{table}

\textbf{Online Evaluation.}
To evaluate temporal grounding when questions arrive during a live video stream, we test CoverFold and TEC using only the video prefix available at query time.
On Charades-STA and ActivityNet, each question is asked at a random time after its target event ends (Random) and again at video end (End).
For longer streams, we evaluate 986 questions over 89 videos of 45--60 minutes from ExtremeWhenBench~\cite{seo2026natural}, querying at different delays after each event and at video end.
All experiments use Qwen3-VL-4B-Instruct at 12.5\% retention, with causal compression and no access to future frames.

\textbf{Real-Time Grounding.}
As shown in \cref{tab:online-grounding}, TEC improves all five pruners on Charades-STA at both query times.
In particular, CoverFold gains 9.53 mIoU at Random, demonstrating that consolidation benefits queries before the stream is complete.
ActivityNet shows more selective improvements, consistent with the smaller gains observed in offline evaluation.

\textbf{Long-Stream Grounding.}
On ExtremeWhenBench, TEC improves the pooled score of every tested pruner, raising CoverFold from 9.41 to 11.71 mIoU (\cref{tab:online-grounding}).
All five pruners also improve within the first five minutes after an event, showing that evidence consolidated before KV eviction remains useful for later queries.

\section{Efficiency}
\label{sec:efficiency-details}
\textbf{Measurement protocol.}
Latency is measured primarily on an RTX 5090 (\cref{tab:stages-5090}); \cref{tab:stages,tab:quantization} provide additional results on an L40 (48\,GB).
Compression and LLM execution are timed together using the measurement scope in \cref{sec:eval-metrics}.
Unless stated otherwise, we use batch-one inference on a one-hour video stream with Qwen3-VL-4B and 12.5\% token retention.
We report mean per-step latency over the final minute of that hour.

\textbf{Stage-wise efficiency.}
To explain the directly measured latency, \cref{tab:stages-5090,tab:stages} provide stage profiles on both platforms.
Each decoder step combines computation---projections, MLPs, and attention---with memory operations: reading weights and historical KV, and copying KV for cache updates and eviction.
We estimate reading costs with a single-new-position probe over the same cache; the excess kernel time estimates computation.
Cache copies are profiled separately, and ``other'' includes small kernels and CPU waits.
Because reading and arithmetic can overlap, these are estimates of their contributions to elapsed time.

TEC increases LLM computation from 904.7 to 4,085.3 TFLOPs, yet adds only 8.9\,ms to the primary platform's decoder step and 15.8\,ms on the additional platform.
The reason is consistent across both: KV reading and copying occupy about 88\% of TEC's decoder time, whereas computation accounts for only 4.5--5.2\,ms.
The extra source tokens share layer weights and largely the same retained history, so their added arithmetic does not multiply the dominant memory traffic.
Thus, a $4.5\times$ operation count produces a much smaller latency increase in this batch-one implementation.

TEC preserves this advantage by keeping the pruner's permanent KV positions unchanged; discarded-token states expire after the short consolidation window.
The cache therefore remains compact instead of retaining every source token for future steps.
This also avoids the host offloading required by the uncompressed one-hour stream in \cref{tab:stages}.
For deployment, the benefit is preserving more evidence within a compact history, rather than minimizing FLOPs alone.

\begin{table}[t]
\caption{\textbf{Stage-wise efficiency on an L40.} $^\dagger$: analytical FLOPs.}
\label{tab:stages}
\centering\footnotesize
\setlength{\tabcolsep}{1.5pt}
\resizebox{1\linewidth}{!}{%
\begin{tabular}{lcccccc}
\toprule
& \multicolumn{2}{c}{\pruner} & \multicolumn{2}{c}{+\method} & \multicolumn{2}{c}{All tokens}\\
\cmidrule(lr){2-3}\cmidrule(lr){4-5}\cmidrule(lr){6-7}
Stage & ms & TFLOPs & ms & TFLOPs & ms & TFLOPs\\
\midrule
Vision encoder & 4.6 & 882.3 & 4.6 & 882.3 & 4.6 & 882.3\\
\pruner selection & 0.78 & 0.16 & 0.78 & 0.16 & -- & --\\
Language model & 169.2 & 904.7 & 185.0 & 4,085.3 & 4,590.5 & 39,819.8$^\dagger$\\
\quad reading weights & 10.7 & -- & 10.7 & -- & 10.7 & --\\
\quad reading KV cache & 81.8 & -- & 82.0 & -- & 687.0 & --\\
\quad KV-cache copies & 69.1 & -- & 80.1 & -- & 3,149.4 & --\\
\quad computing: projections + MLP & 0.8 & 322.7 & 1.6 & 1,452.3 & 2.3 & 2,581.8$^\dagger$\\
\quad computing: attention & 1.1 & 581.9 & 3.6 & 2,633.1 & 0.0 & 37,238.0$^\dagger$\\
\quad other & 5.7 & -- & 7.1 & -- & 741.2 & --\\
\midrule
Total & 174.5 & 1,790.0 & 190.4 & 4,970.2 & 4,595.1 & 40,717.5\\
\bottomrule
\end{tabular}}
\end{table}
\begin{table}[t]
\caption{\textbf{Stage-wise efficiency on an RTX 5090.}}
\label{tab:stages-5090}
\centering\footnotesize
\setlength{\tabcolsep}{3pt}
\begin{tabular}{lcccc}
\toprule
& \multicolumn{2}{c}{\pruner} & \multicolumn{2}{c}{+\method}\\
\cmidrule(lr){2-3}\cmidrule(lr){4-5}
Stage & ms & TFLOPs & ms & TFLOPs\\
\midrule
Vision encoder & 3.4 & 882.3 & 3.4 & 882.3\\
\pruner selection & 1.21 & 0.16 & 1.21 & 0.16\\
Language model & 155.8 & 904.7 & 164.7 & 4,085.3\\
\quad reading weights & 5.4 & -- & 5.2 & --\\
\quad reading KV cache & 104.4 & -- & 107.8 & --\\
\quad KV-cache copies & 34.3 & -- & 37.4 & --\\
\quad computing: projections + MLP & 1.8 & 322.7 & 1.9 & 1,452.3\\
\quad computing: attention & 2.1 & 581.9 & 2.6 & 2,633.1\\
\quad other & 7.8 & -- & 9.8 & --\\
\midrule
Total & 160.5 & 1,790.0 & 169.3 & 4,970.2\\
\bottomrule
\end{tabular}
\end{table}

\begin{table}[t]
\caption{\textbf{Integration with KV quantization on an L40.} W/KV: model weight/KV-cache bits. $^\S$: KV cache offloaded to memory.}
\label{tab:quantization}
\centering\footnotesize
\setlength{\tabcolsep}{2.5pt}
\resizebox{0.75\linewidth}{!}{%
\begin{tabular}{llrrrc}
\toprule
& W/KV & ms & Cache & Total & mIoU \\
\midrule
\multirow{4}{*}{All tokens} & 16/16  & 4595.1$^\S$ & 48.79 & 57.17 & 55.99\\
 & 16/8  & 419.5 & 24.78 & 33.15 & 55.03\\
 & 16/4 & 319.1 & 15.25 & 23.62 & 54.10\\
 & 8/8 & 427.2 & 24.78 & 29.77 & 54.97 \\
\midrule
\multirow{4}{*}{\shortstack[l]{\pruner\\+ \method}} & 16/16  & 190.4 & 6.10 & 14.44 & 53.05 \\
 & 16/8 & 112.8 & 3.10 & 11.43 & 52.98 \\
 & 16/4  & 84.4 & 1.91 & 10.23 & 52.42 \\
 & 8/8  & 113.6 & 3.10 & 8.05 & 52.76 \\
\bottomrule
\end{tabular}}
\end{table}

\textbf{Integration with KV quantization.}
The same bottleneck explains why reducing stored bytes does not automatically reduce latency.
Storage-only quantization reconstructs floating-point KV before passing it to standard PyTorch attention.
Cache concatenation, KV-head expansion, and attention reads can therefore still move 16-bit data, while dequantization adds full-cache reads and writes.
Our low-bit implementation instead stores packed codes and uses a fused Triton kernel to read entries directly and dequantize them in registers.
The kernel supports 16-, 8-, and 4-bit entries; its 16-bit path bypasses dequantization.
This avoids materializing a full floating-point cache, so lower precision can also reduce cache-copy traffic.
Reported latency reflects the complete inference path, including cache handling and attention.

Quantization is a strong baseline: keeping all tokens with 4-bit KV reduces cache size from 48.79 to 15.25\,GiB, avoiding offloading while retaining 54.10 mIoU (\cref{tab:quantization}).
At matched precision---16-bit weights and 4-bit KV---\pruner{} + \method reduces step latency from 319.1 to 84.4\,ms and total memory from 23.62 to 10.23\,GiB compared with retaining all tokens, at a cost of 1.68 mIoU (\cref{tab:quantization}).
The same pattern holds with 8-bit KV.
With \pruner{} + \method, 8-bit KV reduces cache size from 6.10 to 3.10\,GiB while changing mIoU by only $-0.07$ (53.05 to 52.98).
Reducing token count therefore provides substantial savings even after quantization, while TEC helps preserve evidence within the compressed history.

\section{Hybrid-Attention Extension}
\begin{table}[t]
\caption{\textbf{Comparison of TEC state-update strategies on Qwen3.5-9B.}}
\label{tab:hybrid-full}
\centering
\setlength{\tabcolsep}{6pt}
\resizebox{\linewidth}{!}{%
\begin{tabular}{lcccc}
\toprule
& \multicolumn{2}{c}{Charades-STA} & \multicolumn{2}{c}{ActivityNet}\\
\cmidrule(lr){2-3}\cmidrule(lr){4-5}
Qwen3.5-9B, mIoU & 12.5\% & 25\% & 12.5\% & 25\%\\
\midrule
All tokens & \multicolumn{2}{c}{61.25} & \multicolumn{2}{c}{46.34}\\
\midrule
NovaCov & 53.73 & 58.05 & 41.04 & 44.53\\
\midrule
\pruner & 55.01 & 58.54 & 42.06 & 45.50\\
\quad + \method, auxiliary state, layers 1--16 & 56.76 & 59.19 & 40.56 & \textbf{45.70}\\
\quad \textbf{+ \method, main state, layers 1--32 (kept)} & \textbf{58.63} & \textbf{59.67} & \textbf{43.43} & 45.09\\
\bottomrule
\end{tabular}}

\end{table}

\textbf{Adapting TEC to Hybrid Attention.}
Qwen3.5-9B~\cite{qwen35blog} combines 24 Gated DeltaNet linear-attention layers with eight full-attention layers.
Unlike full attention, which stores individually removable token KV, linear attention accumulates history in fixed-size recurrent states, making selective eviction impossible.

We therefore explore two adaptations: \emph{auxiliary-state transfer} lets retained tokens read a separate state containing dropped evidence through layer 16, while preserving only the pruning-only main state; \emph{main-state consolidation} directly incorporates dropped evidence into the persistent state across all 32 layers.
Both retain TEC's $W$-step temporary KV in full-attention layers, but recurrent states may carry dropped evidence beyond this window.
For question answering, ReKV retrieves historical steps in full-attention layers, while linear-attention layers use their accumulated states.

\textbf{Results and Analysis.}
As shown in \cref{tab:hybrid-full}, main-state consolidation improves CoverFold by 3.62 and 1.37 mIoU on Charades-STA and ActivityNet at 12.5\% retention, outperforming auxiliary-state transfer (+1.75 and $-1.50$).
This suggests that directly integrating dropped evidence into the state used by future tokens is more effective than transferring it through a separate, non-persistent pathway.
However, on ActivityNet at 25\%, where pruning already preserves most visual evidence, additional consolidation slightly degrades performance, suggesting interference from redundant information.

Overall, TEC extends to hybrid attention without increasing persistent state size, although its effectiveness depends on how and where discarded evidence is consolidated.

\section{OVO-Bench Evaluation}
\label{sec:ovobench}
\begin{table*}[t]
\caption{\textbf{Online video understanding on OVO-Bench.} Accuracy (\%). Overall averages the three categories; $\Delta$ is its relative change from all tokens. \textbf{Bold} and \underline{underlining} mark the best and second-best compressed results per budget.}
\label{tab:ovobench}
\centering\footnotesize
\setlength{\tabcolsep}{2.4pt}
\resizebox{\linewidth}{!}{%
\begin{tabular}{l*{7}{c}|*{4}{c}|*{4}{c}|cc}
\toprule
& \multicolumn{7}{c|}{Real-Time Visual Perception} & \multicolumn{4}{c|}{Backward Tracing} & \multicolumn{4}{c|}{Forward Active Responding} & & \\
Method & OCR & ACR & ATR & STU & FPD & OJR & Avg & EPM & ASI & HLD & Avg & REC & SSR & CRR & Avg & Overall & $\Delta$\\
\midrule
All tokens & 75.2 & 54.1 & 69.8 & 55.6 & 69.3 & 58.7 & 63.8 & 55.6 & 58.8 & 16.7 & 43.7 & 31.8 & 55.0 & 45.0 & 43.9 & 50.5 & --\\
\midrule
\multicolumn{18}{l}{\textit{Retention 12.5\%}}\\
ToMe~\cite{bolya2022token} & 61.1 & 53.2 & \underline{69.0} & \underline{47.2} & \underline{70.3} & 51.6 & 58.7 & 50.8 & \textbf{58.8} & 14.0 & 41.2 & \underline{19.6} & 48.3 & \textbf{40.4} & 36.1 & 45.4 & $-$10.1\%\\
STC-Pruner~\cite{wang2026accelerating} & \textbf{67.1} & \underline{55.0} & 68.1 & 43.8 & 69.3 & 45.7 & 58.2 & 51.2 & 52.0 & 14.0 & 39.1 & 14.5 & 44.7 & \underline{40.0} & 33.0 & 43.4 & $-$13.9\%\\
NovaCov~\cite{duan2026novacov} & 63.1 & \underline{55.0} & \textbf{71.6} & \textbf{47.8} & 69.3 & \textbf{56.0} & \underline{60.5} & \textbf{52.9} & \underline{57.4} & \underline{15.1} & \textbf{41.8} & 18.3 & \textbf{52.0} & 39.6 & \underline{36.6} & \underline{46.3} & $-$8.3\%\\
\rowcolor{gray!13}\pruner{} + \method (ours) & \underline{63.8} & \textbf{56.0} & \textbf{71.6} & \underline{47.2} & \textbf{74.3} & \underline{53.8} & \textbf{61.1} & \underline{51.5} & 55.4 & \textbf{17.7} & \underline{41.6} & \textbf{27.2} & \underline{50.6} & \underline{40.0} & \textbf{39.3} & \textbf{47.3} & $-$6.3\%\\
\midrule
\multicolumn{18}{l}{\textit{Retention 25\%}}\\
ToMe~\cite{bolya2022token} & \underline{69.8} & \textbf{56.9} & 69.0 & 46.6 & \underline{71.3} & 54.3 & 61.3 & 51.2 & \underline{58.1} & \underline{16.1} & 41.8 & 28.4 & 52.9 & \underline{40.8} & 40.7 & 47.9 & $-$5.0\%\\
STC-Pruner~\cite{wang2026accelerating} & \textbf{73.2} & \textbf{56.9} & 68.1 & 48.9 & 70.3 & 53.3 & 61.8 & \underline{54.9} & \textbf{58.8} & 12.4 & 42.0 & 22.8 & 50.9 & 40.4 & 38.0 & 47.3 & $-$6.3\%\\
NovaCov~\cite{duan2026novacov} & \textbf{73.2} & 55.0 & \underline{69.8} & \underline{49.4} & 68.3 & \textbf{58.2} & \underline{62.3} & 53.9 & \underline{58.1} & 15.1 & \underline{42.3} & \underline{28.5} & \textbf{53.7} & 40.4 & \underline{40.9} & \underline{48.5} & $-$3.9\%\\
\rowcolor{gray!13}\pruner{} + \method (ours) & 67.1 & \underline{56.0} & \textbf{70.7} & \textbf{51.7} & \textbf{72.3} & \underline{57.1} & \textbf{62.5} & \textbf{57.2} & \underline{58.1} & \textbf{17.2} & \textbf{44.2} & \textbf{28.9} & \underline{53.3} & \textbf{42.1} & \textbf{41.4} & \textbf{49.4} & $-$2.2\%\\
\bottomrule
\end{tabular}}
\end{table*}

\textbf{Online protocol.}
We adapt STC's released ReKV evaluation pipeline~\cite{wang2026accelerating,di2025streaming} to Qwen3-VL-4B-Instruct.
Each question is asked at its official evaluation clip's endpoint and sees only already-arrived frames.
We encode each video once and restrict retrieval to steps before each query; causal encoding preserves the memory of that prefix.
Frames are sampled at 0.5\,fps and retain Qwen3-VL's native format of two frames per timestamped step.
We allow up to 392 visual tokens per step, averaging 196 per frame, and retrieve 32 steps (64 frames, about 12.5k tokens), matching ReKV's retrieval size within its 15k-token local window.
All methods use the same backbone, inputs, prompts, retrieval, and retention budgets, differing only in compression.
The 12.5\% and 25\% budgets include timestamp text and are enforced per step without looking ahead.

\textbf{Settings and scoring.}
Baselines use their released defaults.
On this benchmark, \pruner uses change-aware selection and folding with $\gamma=0.4$, $\kappa=0.5$, and $\tau=0$; \method reuses the Video-MME consolidation setting.
We use OVO-Bench's official task prompts and scoring rules~\cite{niu2025ovo}, opening the assistant turn before generation as its chat-model evaluation does.
Task abbreviations follow the benchmark.
Each category averages its task accuracies, and Overall averages the three categories, computed before rounding.

\textbf{Results.}
\pruner{} + \method achieves the highest overall score at both budgets in \cref{tab:ovobench}.
At 12.5\% retention, the clearest gain is in repetition event counting (REC), reaching 27.2 versus 19.6 for the best compared baseline.
Counting requires preserving earlier occurrences as new ones arrive, making a discarded observation potentially relevant to many later answers.
This pattern is consistent with our strategy of consolidating evidence into retained visual and timestamp states before eviction.

At 25\% retention, our approach leads all three category averages, extending this benefit across different online demands.
The gains remain task dependent: backward tracing at 12.5\% stays close to NovaCov, while OCR trails several baselines at 25\%.
Fine spatial detail can remain vulnerable under compression even when event memory improves.
Together, these results support evidence consolidation for online understanding, particularly when past observations must remain useful as the stream evolves.

\end{document}